# A Theoretical Framework for Robustness of (Deep) Classifiers Against Adversarial Examples

**Beilun Wang, Ji Gao, Yanjun Qi**
Department of Computer Science
University of Virginia
Charlottesville, VA 22901, USA
`{bw4mw,jg6yd,yanjun}@virginia.edu`

## Abstract

Most machine learning classifiers, including deep neural networks, are vulnerable to adversarial examples. Such inputs are typically generated by adding small but purposeful modifications that lead to incorrect outputs while imperceptible to human eyes. The goal of this paper is not to introduce a single method, but to make theoretical steps towards fully understanding adversarial examples. By using concepts from topology, our theoretical analysis brings forth the key reasons why an adversarial example can fool a classifier ($f_1$) and adds its oracle ($f_2$, like human eyes) in such analysis. By investigating the topological relationship between two (pseudo)metric spaces corresponding to predictor $f_1$ and oracle $f_2$, we develop necessary and sufficient conditions that can determine if $f_1$ is always robust (strong-robust) against adversarial examples according to $f_2$. Interestingly our theorems indicate that just one unnecessary feature can make $f_1$ not strong-robust, and the right feature representation learning is the key to getting a classifier that is both accurate and strong-robust.

## 1 Introduction

Deep Neural Networks (DNNs) can efficiently learn highly accurate models and have been demonstrated to perform exceptionally well (Krizhevsky et al., 2012; Hannun et al., 2014). However, recent studies show that intelligent attackers can force many machine learning models, including DNNs, to misclassify examples by adding small and hardly visible modifications on a regular test sample.

The maliciously generated inputs are called "adversarial examples" (Goodfellow et al., 2014; Szegedy et al., 2013) and are commonly crafted by carefully searching small perturbations through an optimization procedure. Several recent studies proposed algorithms for solving such optimization to fool DNN classifiers. (Szegedy et al., 2013) firstly observe that convolution DNNs are vulnerable to small artificial perturbations. They use box-constrained Limited-memory BFGS (L-BFGS) to create adversarial examples and find that adversarial perturbations generated from one DNN network can also force other networks to produce wrong outputs. Then, (Goodfellow et al., 2014) try to clarify that the primary cause of such vulnerabilities may be the linear nature of DNNs. They then propose the fast gradient sign method for generating adversarial examples quickly. Subsequent papers (Fawzi et al., 2015; Papernot et al., 2015a; Nguyen et al., 2015) have explored other ways to explore adversarial examples for DNN (details in Section 2.1). The goal of this paper is to analyze the robustness of machine learning models in the face of adversarial examples.

In response to progress in generating adversarial examples, researchers attempt to design strategies for making machine-learning systems robust to various noise, in the worst case as adversarial examples. For instance, denoising NN architectures (Vincent et al., 2008; Gu & Rigazio, 2014; Jin et al., 2015) can discover more robust features by using a noise-corrupted version of inputs as training samples [1]. A modified distillation strategy (Papernot et al., 2015b) is proposed to improve the robustness of DNNs against adversarial examples, though it has been shown to be unsuccessful recently (Carlini & Wagner, 2016a). The most generally successful strategy to date is adversarial training (Goodfellow et al., 2014; Szegedy et al., 2013) which injects adversarial examples into training to improve the generalization of DNN models [2]. More recent techniques incorporate a smoothness penalty (Miyato et al., 2016; Zheng et al., 2016) or a layer-wise penalty (Carlini & Wagner, 2016b) as a regularization term in the loss function to promote the smoothness of the DNN model distributions.

---

[1]Though (Gu & Rigazio, 2014) pointed out such a strategy is vulnerable to attackers who consider the model as a stacked denoiser plus a classifier and compute the gradients of the overall architecture

Table 1: A list of important notations used in the paper.

| $f_1$ | A learned machine learning classifier $f_1 = c_1 \circ g_1$. |
|---|---|
| $f_2$ | The oracle for the same task (see Definition (2.1)) $f_2 = c_2 \circ g_2$. |
| $g_i$ | Part of $f_i$ including operations that progressively transform input into a new form of learned representations in $X_i$. |
| $c_i$ | Part of $f_i$ including simple decision functions (like linear) for classifying. |
| $X$ | Input space (e.g., $\{0, 1, 2, \ldots, 255\}^{32 \times 32 \times 3}$ for CIFAR-10 data (Krizhevsky & Hinton, 2009)). |
| $Y$ | Output space (e.g., $\{1, 2, 3, \ldots, 10\}$ for CIFAR-10 data (Krizhevsky & Hinton, 2009)). |
| $X_1$ | Feature space defined by the feature extraction module $g_1$ of predictor $f_1$. |
| $X_2$ | Feature space defined by the feature extraction module $g_2$ of oracle $f_2$. |
| $d_1(\cdot, \cdot)$ | The metric function for measuring sample distances in feature space $X_1$ with respect to predictor $f_1$. |
| $d_2(\cdot, \cdot)$ | The metric function for measuring sample distance in feature space $X_2$ with respect to oracle $f_2$. |
| $d_1'(\cdot, \cdot)$ | The Pseudometric function with respect to predictor $f_1$, $d_1'(x, x') = d_1(g_1(x), g_1(x'))$. |
| $d_2'(\cdot, \cdot)$ | The Pseudometric function with respect to oracle $f_2$, $d_2'(x, x') = d_2(g_2(x), g_2(x'))$. |
| $a.e.$ | almost everywhere (Folland, 2013); (defined by Definition (8.2) in Section 8.1) |
| $\epsilon, \delta_1, \delta_2, \delta, \eta$ | small positive constants |

Recent studies (reviewed by (Papernot et al., 2016b)) are mostly empirical and provide little understanding of why an adversary can fool machine learning models with adversarial examples. Several important questions have not been answered yet:

- What makes a classifier always robust to adversarial examples?
- Which parts of a classifier influence its robustness against adversarial examples more, compared with the rest?
- What is the relationship between a classifier's generalization accuracy and its robustness against adversarial examples?
- Why (many) DNN classifiers are not robust against adversarial examples ? How to improve?

This paper tries to answer above questions and makes the following contributions:

- Section 2 points out that previous definitions of adversarial examples for a classifier ($f_1$) have overlooked the importance of an oracle function ($f_2$) of the same task.
- Section 4 formally defines when a classifier $f_1$ is always robust ("strong-robust") against adversarial examples. It proves four theorems about sufficient and necessary conditions that make $f_1$ always robust against adversarial examples according to $f_2$. Our theorems lead to a number of interesting insights in Section 5, like that the feature representation learning controls if a DNN is strong-robust or not.
- Section 6 is dedicated to provide practical and theoretically grounded directions for understanding and hardening DNN models against adversarial examples.

Table 1 provides a list of important notations we use in the paper.

---

[2] Though (Papernot et al., 2016b) pointed out that adversarial training is vulnerable to black-box attacks using model transferability

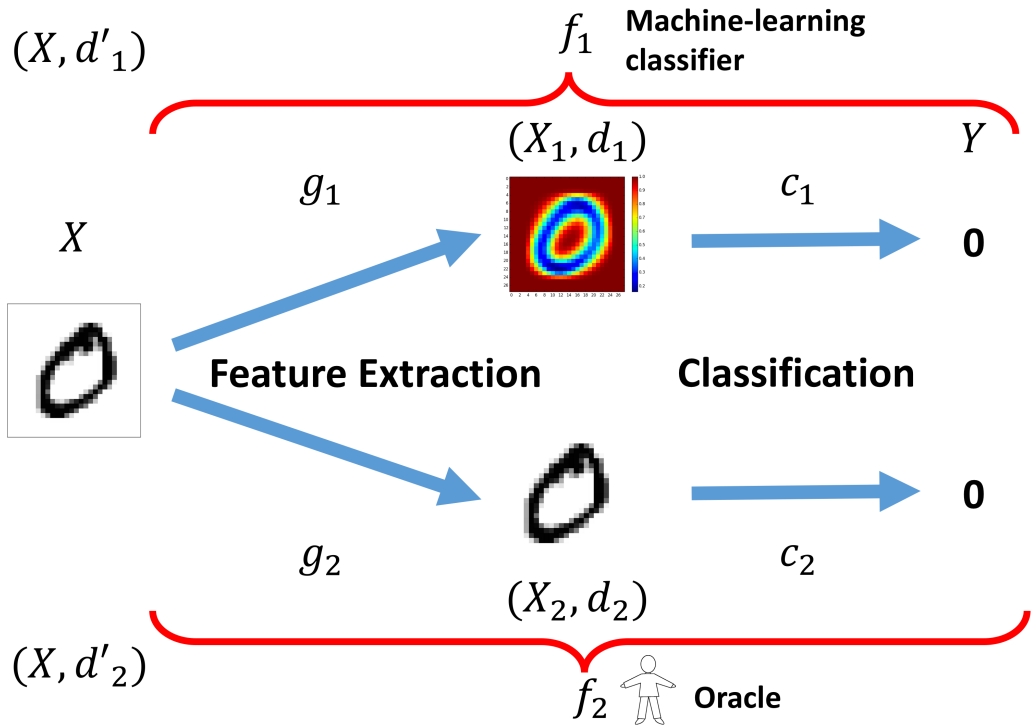

Figure 1: Example of a machine-learning classifier (predictor) and a human annotator (oracle) for classifying images of hand-written "0". Both include two steps: feature extraction and classification. The upper half is about the learned machine classifier $f_1$ and the lower half is about the oracle $f_2$. $f_1$ transforms samples from the original space $X$ to an embedded metric space $(X_1, d_1)$ using its feature extraction step. Here, $d_1$ is the similarity measure on the feature space $X_1$. Classification models like DNN cover the feature extraction step in their model, though many other models like decision tree need hard-crafted or domain-specific feature extraction. Then $f_1$ can use a linear function to decide the classification prediction $\hat{y} \in Y$. Similarly, human oracle $f_2$ transforms data samples from the original space $X$ into an embedded metric space $(X_2, d_2)$ by its feature extraction. Here, $d_2$ is the corresponding similarity measure. Then the oracle get the classification result $y \in Y$ using the feature representation of samples $(X_2, d_2)$.

## 2 Re-Define Adversarial Examples

This section provides a general definition of *adversarial examples* , by including the notion of an oracle. For a particular classification task, a learned classifier is represented as $f_1 : X \rightarrow Y$, where $X$ represents the input sample space and $Y$ is the output space representing a categorical set.

### 2.1 Previous Formulations

Various definitions of "adversarial examples" exist in the recent literature, with most following Eq. (2.1). See more detailed reviews in Section 9. The basic idea is to generate a misclassified sample $x'$ by "slightly" perturbing a correctly classified sample $x$, with an adversarial perturbation $\Delta(x, x')$. Formally, when given $x \in X$

$$\text{Find } x'$$
$$\text{s.t. } f_1(x) \neq f_1(x') \tag{2.1}$$
$$\Delta(x, x') < \epsilon$$

Here $x, x' \in X$. $\Delta(x, x')$ represents the difference between $x$ and $x'$, which depends on the specific data type that $x$ and $x'$ belong to [3]. Table 2 summarizes different choices of $f_1$ and $\Delta(x, x')$ used in the recent literature, in which norm functions on the original space $X$ are mostly used to calculate $\Delta(x, x')$. Multiple algorithms have been implemented to solve Eq. (2.1) as a constrained optimization (summarized by the last column of Table 2). More details are included for three such studies in Section 3.2.

---

[3]For example, in the case of strings, $\Delta(x, x')$ represents the difference between two strings.

When searching for adversarial examples, one important property has not been fully captured by Eq. (2.1). That is, an adversarial example has been modified very slightly from its seed and these modifications can be so subtle that, for example in image classification, a human observer does not even notice the modification at all. We define the role of "human observer" more formally as follows:

**Definition 2.1.** *An "**Oracle**" represents a decision process generating ground truth labels for a task of interest. Each oracle is task-specific, with finite knowledge and noise-free*[4].

The goal of machine learning is to train a learning-based predictor function $f_1 : X \to Y$ to approximate an oracle classifier $f_2 : X \to Y$ for the same classification task. For example, in image classification tasks, the oracle $f_2$ is often a group of human annotators. Adding the notation of oracle, we revise Eq. (2.1) into:

$$\begin{aligned}
&\text{Find } x' \\
&\text{s.t. } f_1(x) \neq f_1(x') \\
&\quad \Delta_2(x, x') < \epsilon \\
&\quad f_2(x) = f_2(x')
\end{aligned} \qquad (2.2)$$

## 2.2 Measuring Sample Difference in Which Space? Modeling & Decomposing $f_2$

$\Delta_2(x, x') < \epsilon$ reflects that adversarial examples add "small modifications" that are almost imperceptible to oracle of the task. Clearly calculating $\Delta_2(x, x')$ needs to accord to oracle $f_2$. For most classification tasks, an oracle does not measure the sample difference in the original input space $X$. We want to emphasize that sample difference is with regards to its classification purpose. For instance, when labeling images for the hand-written digital recognition, human annotators do not need to consider those background pixels to decide if an image is "0" or not.

Illustrated in Figure 1, we denote the feature space an oracle uses to consider difference among samples for the purpose of classification decision as $X_2$. The sample difference uses a distance function $d_2$ in this space. An oracle function $f_2 : X \to Y$ can be decomposed as $f_2 = c_2 \circ g_2$ where $g_2 : X \to X_2$ represents the operations for feature extraction from $X$ to $X_2$ and $c_2 : X_2 \to Y$ denotes the simple operation of classification in $X_2$. Essentially $g_2$ includes the operations that (progressively) transform input representations into an informative form of representations $X_2$. $c_2$ applies relatively simple functions (like linear) in $X_2$ for the purpose of classification. $d_2$ is the metric function (details in Section 4) an oracle uses to measure the similarity among samples (by relying on representations learned in the space $X_2$). We illustrate the modeling and decomposition in Figure 1.

In Section 4 our theoretical analysis uses $(X_2, d_2)$ to bring forth the fundamental causes of adversarial examples and leads to a set of novel insights to understand such examples. To the best of the authors' knowledge, the theoretical analysis made by this paper has not been uncovered by the literature.

**Modeling Oracle $f_2$:** One may argue that it is hard to model $f_2$ and $(X_2, d_2)$ for real applications, since if such oracles can be easily modeled machine-learning based $f_1$ seems not necessary. In Section 9.1, we provide examples of modeling oracles for real applications. For many security-sensitive applications about machines, oracles $f_2$ do exist [5]. For artificial intelligence tasks like image classification, humans are $f_2$. As illustrated by cognitive neuroscience papers (DiCarlo & Cox, 2007; DiCarlo et al., 2012), human brains perform visual object recognition using the ventral visual stream, and this stream is considered to be a progressive series of visual re-representations, from V1 to V2 to V4 to IT cortex (DiCarlo & Cox, 2007). Experimental results support that human visual system makes classification decision at the final IT cortex layer. This process is captured exactly by our decomposition $f_2 = c_2 \circ g_2$.

---

[4]We leave all detailed analysis of when an oracle contains noise as future work.

[5]Oracles $f_2$ do exist in many security-sensitive applications about machines. But machine-learning classifiers $f_1$ are used popularly due to speed or efficiency

Table 2: Summary of the previous studies defining adversarial examples.

| Previous studies | $f_1$ | $\Delta(x, x')$ | Formulation of $f_1(x) \neq f_1(x')$ |
|---|---|---|---|
| (Goodfellow et al., 2014) | Convolutional neural networks | $\ell_\infty$ | $\underset{x'}{\text{argmax}}\, Loss(f_1(x'), f_1(x))$ |
| (Papernot et al., 2016a) | Convolutional neural networks | $\ell_0$ | $\underset{x'}{\text{argmax}}\, Loss(f_1(x'), f_1(x))$ |
| (Grosse et al., 2016) | Convolutional neural networks | $\ell_0$ | $\underset{x'}{\text{argmax}}\, Loss(f_1(x'), f_1(x))$ |
| (Szegedy et al., 2013) | Convolutional neural networks | $\ell_2$ | $\underset{x'}{\text{argmin}}\, Loss(f_1(x'), l)$, when: $f_1(x) \neq l$ |
| (Biggio et al., 2013) | Support vector machine (SVM) | $\ell_2$ | $\underset{x'}{\text{argmin}}\, Loss(f_1(x'), -1)$, when: $f_1(x) = 1$ |
| (Kantchelian et al., 2015) | Decision tree and Random forest | $\ell_2$, $\ell_1$, $\ell_\infty$ | $\underset{x'}{\text{argmin}}\, Loss(f_1(x'), -1)$, when: $f_1(x) = 1$ |
| (Xu et al., 2016) | Random forest and SVM | $\ell_1, \ell_\infty$ | $\underset{x'}{\text{argmin}}\, Loss(f_1(x'), -1)$, when: $f_1(x) = 1$ |

## 2.3 Revised Formulation

Now we use the decomposition of $f_2$ to rewrite $\Delta_2(x, x')$ as $d_2(g_2(x), g_2(x'))$ in Eq. (2.2) and obtain our proposed general definition of adversarial examples:

**Definition 2.2.** *adversarial example: Suppose we have two functions $f_1$ and $f_2$. $f_1 : X \to Y$ is the classification function learned from a training set and $f_2 : X \to Y$ is the classification function of the oracle that generates ground-truth labels for the same task. Given a sample $x \in X$, an adversarial example $x' \in X$. $(x, x')$ satisfies Eq. (2.3).*

$$
\begin{aligned}
&\textit{Find } x' \\
&\textit{s.t. } f_1(x) \neq f_1(x') \\
&d_2(g_2(x), g_2(x')) < \delta_2 \\
&f_2(x) = f_2(x')
\end{aligned}
\quad (2.3)
$$

Most previous studies (Table 2) have made an important and implicit assumption about $f_2$ (through using $\Delta(x, x') < \epsilon$): $f_2$ is almost everywhere (a.e.) continuous. We explains the a.e. continuity assumption and its indication in Section 8. Basically, when $f_2$ is assumed continuous a.e.,

$$
\mathbb{P}(f_2(x) = f_2(x') | d_2(g_2(x), g_2(x')) < \delta_2) = 1
$$

Therefore, when $f_2$ is continuous a.e. Eq. (2.3) can be simplified into the following Eq. (2.4).

$$
\begin{aligned}
&\text{Find } x' \\
&\text{s.t. } f_1(x) \neq f_1(x') \\
&d_2(g_2(x), g_2(x')) < \delta_2
\end{aligned}
\quad (2.4)
$$

# 3 Background: Adversarial Examples

## 3.1 Background: Previous Definitions of Adversarial Examples

For the purpose of "fooling" a classifier, naturally, the attacker wants to control the size of the perturbation $\Delta(x, x')$ to ensure the perturbed sample $x'$ still stays close enough to the original sample $x$ to satisfy the intended "fooling" purpose. For example, in the image classification case, Eq. (2.1) can use the gradient information to find a $\Delta(x, x')$ that makes *human annotators* still recognize $x'$ as almost the same as $x$, though the classifier will predict $x'$ into a different class. In another example with more obvious security implications about PDF malware (Xu et al., 2016), $x'$ in Eq. (2.1) is found by genetic programming. A modified PDF file from a malicious PDF seed will still be recognized as malicious by *an oracle machine* (i.e., a virtual machine decides if a PDF file is malicious or not by actually running it), but are classified as benign by state-of-art machine learning classifiers (Xu et al., 2016).

The following Eq. (3.1) has been popular as well. Eq. (3.1) is a special case of Eq. (2.1).

$$\underset{x' \in X}{\operatorname{argmin}} \ \Delta(x, x')$$

$$\text{Subject to: } f_1(x) \neq f_1(x') \tag{3.1}$$

Eq. (3.1) tries to find the $x'$ by minimizing $\Delta(x, x')$ under some constraints. Eq. (2.1) is a more general formulation than Eq. (3.1) and can summarize most relevant studies. For example, in (Xu et al., 2016) "adversarial examples" are those generated PDFs that can fool PDFRate (a learning-based classifier for detecting malicious PDFs) to classify them as benign. The distances of these variant PDFs to the seed PDF are not necessarily minimal. For such cases, Eq. (2.1) still fits, while Eq. (3.1) does not.

Besides, in the field of computer security, machine learning has been popular in classifying the malicious ($y = 1$) behavior versus benign behavior ($y = -1$). For such a context, two different definitions of adversarial examples exist in the literature:

For instance, (Biggio et al., 2013) uses a formula as follows:

$$\underset{x'}{\operatorname{argmin}}(f_1(x'))$$

$$\text{s.t. } \Delta(x, x') < d_{\max} \tag{3.2}$$

$$f_1(x) > 0$$

Differently, (Lowd & Meek, 2005) uses the following formula:

$$\underset{x'}{\operatorname{argmin}}(\Delta(x, x'))$$

$$\text{s.t. } f_1(x') < 0 \tag{3.3}$$

$$f_1(x) > 0$$

Here $d_{\max}$ is a small positive constant. These definitions of "adversarial examples" are special cases of Eq. (3.1) and Eq. (2.1).

### 3.2 Background: Previous Algorithms Generating "adversarial examples"

To fool classifiers at test time, several approaches have been implemented to generate "adversarial perturbations" by solving Eq. (2.2). According to Eq. (2.2), an adversarial example should be able to change the classification result $f_1(x)$, which is a discrete value. To solve Eq. (2.2), we need to transform the constraint $f_1(x) \neq f_1(x')$ into an optimizable formulation. Then we can easily use the Lagrangian multiplier to solve Eq. (2.2). All the previous studies define a loss function $Loss(\cdot, \cdot)$ to quantify the constraint $f_1(x) \neq f_1(x')$. This loss function can be the same with the training loss, or it can be chosen differently, such as hinge loss or cross entropy loss.

We summarize four common attacking studies as follows:

**Gradient ascent method (Biggio et al., 2013)**   Machine learning has been popular in classifying malicious ($y = 1$) versus benign ($y = -1$) in computer security tasks. For such contexts, a simple way to solve Eq. (2.2) is through gradient ascent. To minimize the size of the perturbation and maximize the adversarial effect, the perturbation should follow the gradient direction (i.e., the direction providing the largest increase of function value, here from $y = -1$ to 1). Therefore, the perturbation $r$ in each iteration is calculated as:

$$r = \epsilon \nabla_x Loss(f_1(x + r), 1) \text{ when: } f_1(x) = 1 \tag{3.4}$$

By varying $\epsilon$, this method can find a sample $x'$ with regard to $d_2(x, x')$ such that $f_1(x) \neq f_1(x')$.

**Box L-BFGS adversary (Szegedy et al., 2013)**   This study views the adversarial problem as a constrained optimization problem, i.e., find a minimum perturbation in the restricted sample space. The perturbation is obtained by using Box-constrained L-BFGS to solve the following equation (reformulated by the method of Lagrange multipliers ):

$$\underset{r}{\operatorname{argmin}}(c \times d_2(x, x + r) + Loss(f_1(x + r), l)), x + r \in [0, 1]^p \tag{3.5}$$

Here $p$ is the total number of features, $c$ is a term added for the Lagrange multiplier. (for an image classification task, it is 3 times the total number of pixels of an RGB image) $l$ is a target label, which is different from the original label. The constraint $x + r \in [0, 1]^p$ means that the adversarial example is still in the range of sample space.

**Fast gradient sign method (Goodfellow et al., 2014)**  The fast gradient sign method proposed by (Goodfellow et al., 2014) views $d_2$ as the $\ell_\infty$-norm. In this case, a natural choice is to make the attack strength at every feature dimension the same. The perturbation is obtained by solving the following equation (reformulated by the method of Lagrange multipliers ):

$$\underset{r}{\operatorname{argmin}}(c \times d_2(x, x+r) - Loss(f_1(x+r), f_1(x))), x + r \in [0, 1]^p \qquad (3.6)$$

Therefore the perturbation can be calculated directly by:

$$r = \epsilon \operatorname{sign}(\nabla_x Loss(f_1(x+r), f_1(x))) \qquad (3.7)$$

Here the loss function is the function used to train the neural network. A recent paper (Kurakin et al., 2016) shows that adversarial examples generated by fast gradient sign method are misclassified even after these images have been recaptured by cameras.

**Jacobian-based saliency map approach (Papernot et al., 2015a)**  (Papernot et al., 2015a) proposed the Jacobian-based saliency map approach (JSMA) to search for adversarial samples while limiting the number of pixel in the image. As a targeted attack, JSMA iteratively perturbs those pixels in an input that provide larger values in the adversarial saliency scores. The adversarial saliency map is calculated from the Jacobian (gradient) matrix $\nabla_x f_1(\mathbf{x})$ of the DNN model at the current input $\mathbf{x}$. The $(i, j)^{\text{th}}$ component in Jacobian matrix $\nabla_x f_1(\mathbf{x})$ describes the derivative of output class $j$ with respect to the pixel $i$. For each pixel $i$, its adversarial saliency score is calculated to reflect how this pixel will increase the output score of class $j$ versus changing the score of other possible output classes. The process is repeated until misclassification in the target class is achieved or the maximum number of perturbed pixels has been reached. Essentially, JSMA optimizes Equation 2.1 by measuring perturbation $\Delta(\mathbf{x}, \mathbf{x}')$ through the $\ell_0$-norm.

### 3.3 BACKGROUND: RELATED WORKS IN A BROADER CONTEXT

Investigating the behavior of machine learning systems in adversarial environments is an emerging topic (Huang et al., 2011; Barreno et al., 2006; 2010; Globerson & Roweis, 2006; Biggio et al., 2013; Kantchelian et al., 2015; Zhang et al., 2015). Recent studies can be roughly categorized into three types: (1) Poisoning attacks in which specially crafted attack points are injected into the training data. Multiple recent papers (Alfeld et al., 2016; Mei & Zhu, 2015b; Biggio et al., 2014; 2012; Mei & Zhu, 2015a) have considered the problem of an adversary being able to pollute the training data with the goal of influencing learning systems including support vector machines (SVM), autoregressive models and topic models. (2) Evasion attacks are attacks in which the adversary's goal is to create inputs that are misclassified by a deployed target classifier. Related studies (Szegedy et al., 2013; Goodfellow et al., 2014; Xu et al., 2016; Kantchelian et al., 2015; Rndic & Laskov, 2014; Biggio et al., 2013; Papernot et al., 2016b; Sinha et al., 2016) assume the adversary does not have an opportunity to influence the training data, but instead finds "adversarial examples" to evade a trained classifier like DNN, SVM or random forest. (3) Privacy-aware machine learning (Duchi et al., 2014) is another important category relevant to data security in machine learning systems. Recent studies have proposed various strategies (Xie et al., 2014; Bojarski et al., 2014; Stoddard et al., 2014; Li & Zhou, 2015; Rajkumar & Agarwal, 2012; Dwork, 2011; Nock et al., 2015) to preserve the privacy of data such as differential privacy. This paper focuses on evasion attacks that are mostly used to attacking classifiers that try to distinguish malicious behaviors from benign behaviors. Here we extend it to a broader meaning – adversarial manipulation of test samples. Evasion attacks may be encountered during system deployment of machine learning methods in adversarial settings.

In the broader secure machine learning field, researchers also make attempts for hardening learning systems. For instance: (1) (Barreno et al., 2010) and (Biggio et al., 2008) propose a method to introduce some randomness in the selection of classification boundaries; (2) A few recent studies (Xiao et al., 2015; Zhang et al., 2015) consider the impact of using reduced feature sets on classifiers under adversarial attacks. (Xiao et al., 2015) proposes an adversary-aware feature selection model that can improve a classifier's robustness against adversarial attacks by incorporating specific assumptions about the adversary's data manipulation strategy. (3) Another line of works, named as adversarial training (Goodfellow et al., 2014), designs a new loss function for training neural networks, which is a linear interpolation of the loss function of the original sample and the loss function of the adversarial example generated by the original sample. A scalable version of adversarial training (Kurakin et al., 2016) was recently proposed. By applying several tricks, the author can apply the adversarial training to deeper network trained by the imagenet dataset. (4) Multiple studies model adversarial scenarios with formal frameworks representing the interaction between the classifier and the adversary. Related efforts include perfect information assumptions (Dalvi et al., 2004), assuming a polynomial number of membership queries (Lowd & Meek, 2005), formalizing the attack process as a two-person sequential Stackelberg game (Brückner & Scheffer, 2011; Liu & Chawla, 2010), a min-max strategy (training a classifier with best performance under the worst perturbation) (Dekel et al., 2010; Globerson & Roweis, 2006), exploring online and non-stationary learning (Dahlhaus, 1997; Cesa-Bianchi & Lugosi, 2006), and formalizing as an adversarial reinforcement learning problem (Uther & Veloso, 1997). (5) A PAC model study about learning adversary behavior in a security games also investigated the solution of computing the best defender strategy against the learned adversary behavior. It has a similar conclusion as ours ( Section 4) that the extreme cases that the defender doesn't work only has zero probability (Sinha et al., 2016).

## 4 DEFINE STRONG-ROBUSTNESS

With a more accurate definition of "adversarial examples", now we aim to answer the first central question: "What makes a classifier always robust against adversarial examples?". Section 4.2 defines the concept "strong-robust" describing a classifier always robust against adversarial examples. Section 4.3 and Section 4.4 present sufficient and necessary conditions for "strong-robustness". Section 5 then provides a set of theoretical insights to understand "strong-robustness".

### 4.1 MODELING AND DECOMPOSING $f_1$

As shown in Figure 1, we decompose $f_1$ in a similar way as the decomposition of $f_2$. This is to answer another key question: "which parts of a learned classifier influence its robustness against adversarial examples more, compared with the rest?". A machine-learning classifier $f_1 = c_1 \circ g_1$, where $g_1 : X \to X_1$ represents the feature extraction operations and $c_1 : X_1 \to Y$ performs a simple operation (e.g., linear) of classification. Section 9.2 provides multiple examples of decomposing state-of-the-art $f_1$ [6]. $d_1$ denotes the distance function $f_1$ uses to measure difference among samples in $X_1$.

Almost all popular machine learning classifiers satisfy the a.e. continuity assumption. It means:
$$\mathbb{P}(f_1(x) = f_1(x')|d_1(g_1(x), g_1(x')) < \delta_1) = 1$$
When $f_1$ is not continuous a.e., it is not robust to any types of noise. See Section 8 for detailed discussions.

For the rare cases that $f_1$ is not continuous a.e., Section 13 discusses "boundary points" of $f_1$ [7]. Roughly speaking, when $f_1$ is not continuous a.e., [8]
$$\mathbb{P}(f_1(x) \neq f_1(x')|d_1(g_1(x), g_1(x')) < \delta_1) > 0$$
Therefore the following probability of "boundary points based adversarial examples" might not be 0 for such cases [9]:
$$\begin{aligned} \mathbb{P}(f_1(x) \neq f_1(x')|f_2(x) = f_2(x'), \\ d_1(g_1(x), g_1(x')) < \delta_1, d_2(g_2(x), g_2(x')) < \delta_2) \end{aligned} \quad (4.1)$$
The value of this probability is critical for our analysis in Theorem (4.3) and in Theorem (4.5).

### 4.2 $\{\delta_2, \eta\}$-STRONG-ROBUST AGAINST ADVERSARIAL EXAMPLES

We then apply reverse-thinking on Definition (2.2) and derive the following definition of **strong-robustness** for a machine learning classifier against adversarial examples:

**Definition 4.1.** $\{\delta_2, \eta\}$-***Strong-robustness of a machine-learning classifier:*** *A machine-learning classifier $f_1(\cdot)$ is $\{\delta_2, \eta\}$-strong-robust against adversarial examples if: $\forall x, x' \in X$ a.e., $(x, x')$ satisfies Eq. (4.2).*

---

[6] Notice that $g_1$ may also include implicit feature selection steps like $\ell_1$ regularization.

[7] Boundary points are those points satisfying $f_1(x) \neq f_1(x')$ and $d_1(g_1(x), g_1(x')) < \delta_1$

[8] When $f_1$ is continuous a.e., $\mathbb{P}(f_1(x) \neq f_1(x')|d_1(g_1(x), g_1(x')) < \delta_1) = 0.$

[9] "Boundary points based adversarial examples" only attack seed samples who are boundary points of $f_1$.

$$\forall x, x' \in X$$
$$\mathbb{P}(f_1(x) = f_1(x') | f_2(x) = f_2(x'),$$
$$d_2(g_2(x), g_2(x')) < \delta_2) > 1 - \eta \tag{4.2}$$

When $f_2$ is continuous a.e., Eq. (4.2) simplifies into Eq. (4.3):
$$\forall x, x' \in X, \ \mathbb{P}(f_1(x) = f_1(x') |$$
$$d_2(g_2(x), g_2(x')) < \delta_2) > 1 - \eta \tag{4.3}$$

Eq. (4.2) defines the "$\{\delta_2, \eta\}$-strong-robustness" as a claim with the high probability. To simplify notations, in the rest of this paper, we use "strong-robust" representing "$\{\delta_2, \eta\}$-strong-robust". Also in the rest of this paper we propose and prove theorems and corollaries by using its more general form by Eq. (4.2). For all cases, if $f_2$ is continuous a.e., all proofs and equations can be simplified by using only the term $d_2(g_2(x), g_2(x')) < \delta_2$ (i.e. removing the term $f_2(x) = f_2(x')$) according to Eq. (4.3)).

The "strong-robustness" definition leads to four important theorems in next two subsections.

### 4.3 Topological Equivalence of Two Metric Spaces $(X_1, d_1)$ and $(X_2, d_2)$ is Sufficient in Determining Strong-robustness

In the appendix, Section 10.1 briefly introduces the concept of metric space and the definition of topological equivalence among two metric spaces. As shown in Figure 1, here $f_1$ defines a metric space $(X_1, d_1)$ on $X_1$ with the metric function $d_1$. Similarly $f_2$ defines a metric space $(X_2, d_2)$ on $X_2$ with the metric function $d_2$.

If the topological equivalence ( Eq. (10.1)) exists between $(X_1, d_1)$ and $(X_2, d_2)$, it means that for all pair of samples from $X$, we have the following relationship:
$$\forall x, x' \in X,$$
$$d_1(g_1(x), g_1(x')) < \delta_1 \Leftrightarrow d_2(g_2(x), g_2(x')) < \delta_2 \tag{4.4}$$

When $f_1$ is continuous a.e., this can get us the following important theorem, indicating that the topological equivalence between $(X_1, d_1)$ and $(X_2, d_2)$ is a sufficient condition in determining whether or not $f_1$ is strong-robust against adversarial examples:

**Theorem 4.2.** *When $f_1$ is continuous a.e., if $(X_1, d_1)$ and $(X_2, d_2)$ are topologically equivalent, then the learned classifier $f_1(\cdot)$ is strong-robust to adversarial examples.*

*Proof.* See its proofs in Section 10.3.4 □

This theorem can actually guarantee that:
$$\forall x, x' \in X,$$
$$\mathbb{P}(f_1(x) = f_1(x') | f_2(x) = f_2(x'),$$
$$d_2(g_2(x), g_2(x')) < \delta_2) = 1 \tag{4.5}$$
Clearly Eq. (4.5) is a special (stronger) case of the "strong-robustness" defined by Eq. (4.2).

For more general cases including $f_1$ might not be continuous a.e., we need to consider the probability of the boundary point attacks (Eq. (4.1)). Therefore, we get a more general theorem as follows:

**Theorem 4.3.** *If $(X_1, d_1)$ and $(X_2, d_2)$ are topologically equivalent and $\mathbb{P}(f_1(x) \neq f_1(x') | f_2(x) = f_2(x'), d_1(g_1(x), g_1(x')) < \delta_1, d_2(g_2(x), g_2(x')) < \delta_2) < \eta$, then the learned classifier $f_1(\cdot)$ is strong-robust to adversarial examples.*

*Proof.* See its proofs in Section 10.3.3. □

### 4.4 Finer Topology of $(X, d'_1)$ than $(X, d'_2)$ is Sufficient and Necessary in Determining Strong-robustness

Now we extend the discussion from two metric spaces into two pseudometric spaces. This extension finds the sufficient and necessary condition that determines the strong-robustness of $f_1$. The related two pseudometrics are $d_1'$ (for $f_1$) and $d_2'$ (for $f_2$), both directly being defined on $X$. Appendix Section 10.2 includes detailed descriptions of pseudometric, pseudometric spaces, topology and a finer topology relationship between two pseudometric spaces.

Table 3: Summary of theoretical conclusions that we can derive. Here $X_1 = \mathbb{R}^{n_1}$ and $X_2 = \mathbb{R}^{n_2}$. The strong-robustness is determined by feature extraction function $g_1$. The accuracy is determined by both the classification function $c_1$ and the feature extraction function $g_1$.

| | Cases: | $d_1 \& d_2$ are **norms** | Can be accurate? | Based on | Illustration |
|---|---|---|---|---|---|
| (I) | $X_1 \setminus (X_1 \bigcap X_2) \neq \emptyset$, $X_2 \not\subset X_1$ | **Not Strong-robust** | can not be accurate | Theorem (4.4) | Partially by Fig. 2 |
| (II) | $n_1 > n_2, X_2 \subsetneq X_1$ | **Not strong-robust** | can be accurate | Corollary (5.1) | Figure 2 |
| (III) | $n_1 = n_2, X_1 = X_2$ | **Strong-robust** | can be accurate | Corollary (5.2) | Figure 4 |
| (IV) | $n_1 < n_2, X_1 \subset X_2$ | **Strong-robust** | can not be accurate | Theorem (4.4) | Figure 5 |

Essentially, the topology in pseudometric space $(X, d_1')$ is a finer topology than the topology in pseudometric space $(X, d_2')$ means:

$$\forall x, x' \in X, d_2'(x, x') < \delta_2 \Rightarrow d_1'(x, x') < \delta_1 \qquad (4.6)$$

Because $d_1'(x, x') = d_1(g_1(x), g_1(x'))$ and $d_2'(x, x') = d_2(g_2(x), g_2(x'))$, the above equation equals to:

$$\begin{aligned} &\forall x, x' \in X, \\ &d_2(g_2(x), g_2(x')) < \delta_2 \Rightarrow d_1(g_1(x), g_1(x')) < \delta_1 \end{aligned} \qquad (4.7)$$

Using Eq. (4.7) and the continuity a.e. assumption, we can derive the following Theorem about the sufficient and necessary condition for $f_1$ being strong-robust:

**Theorem 4.4.** *When $f_1$ is continuous a.e., $f_1$ is strong-robust against adversarial examples if and only if the topology in $(X, d_1')$ is a finer topology than the topology in $(X, d_2')$.*

*Proof.* See its proof in appendix Section 10.3.1. □

Actually the above theorem can guarantee that when $f_1$ is continuous a.e.:

$$\forall x, x' \in X, \mathbb{P}(f_1(x) = f_1(x') | d_2(g_2(x), g_2(x')) < \delta_2) = 1 \qquad (4.8)$$

Eq. (4.8) clearly is a special (stronger) case of strong-robustness defined by Eq. (4.2).

When $f_1$ is not continuous a.e., we need to consider the probability of the boundary points based adversarial examples (Eq. (4.1)). For such a case, we get a sufficient condition [10] for the strong-robustness:

**Theorem 4.5.** *When $f_1$ is not continuous a.e., if the topology in $(X, d_1')$ is a finer topology than the topology in $(X, d_2')$ and $\mathbb{P}(f_1(x) \neq f_1(x') | f_2(x) = f_2(x'), d_1(g_1(x), g_1(x')) < \delta_1, d_2(g_2(x), g_2(x')) < \delta_2) < \eta$, then $f_1$ is strong-robust against adversarial examples.*

When $f_1$ is not continuous a.e., its strong-robustness is significantly influenced by its boundary points and therefore relates to the $c_1$ function. Section 13.2 provides some discussion and we omit covering such cases in the rest of this paper.

## 5 Towards Principled Understanding

The four theorems proposed above lead to a set of key insights about why and how an adversarial can fool a machine-learning classifier using adversarial examples. One of the most valuable insights is: feature learning step decides whether a predictor is strong-robust or not in an adversarial test setting. All the discussions in the subsection assume $f_1$ is continuous a.e..

---

[10]When $f_1$ is not continuous a.e., it is difficult to find the necessary and sufficient condition for strong-robustness of $f_1$. We leave this to future research.

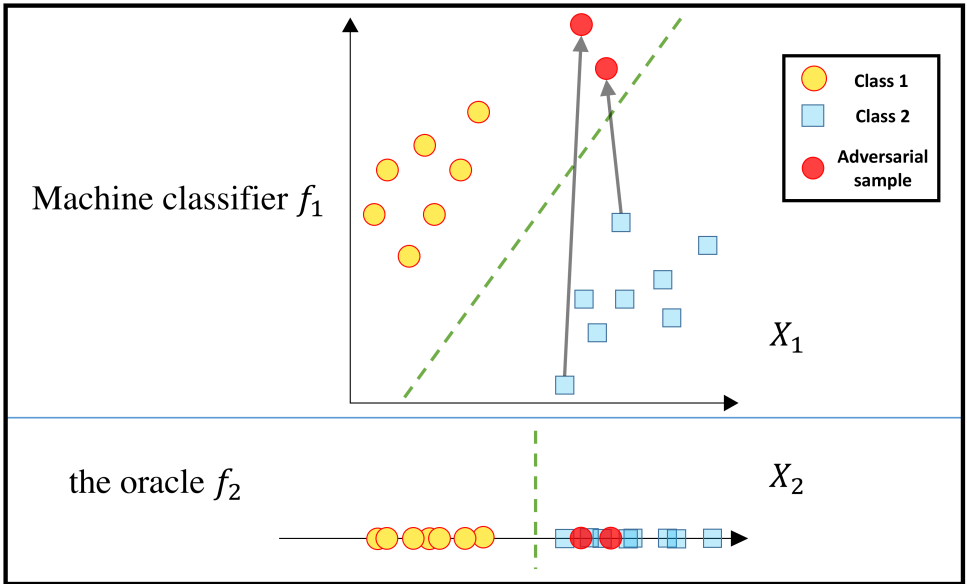

Figure 2: An example showing that $f_1$ with one unnecessary feature (according to $f_2$) is prone to adversarial examples. The red circle denotes an adversarial example (e.g. generated by some attack similar as JSMA (Papernot et al., 2015a) (details in Section 3.2)). Each adversarial example is very close to its seed sample in the oracle feature space (according to $d_2$), but it is comparatively far from its seed sample in the feature space (according to $d_1$) of the trained classifier and is at the different side of the decision boundary of $f_1$. Essentially "adversarial examples" can be easily found for all seed samples in this Figure. We only draw cases for two seeds. Besides, for each seed sample, we can generate a series of "adversarial examples" (by varying attacking power) after the attacking line crosses the decision boundary of $f_1$. We only show one case of such an adversarial example for each seed sample.

### 5.1 UNNECESSARY FEATURES RUIN STRONG-ROBUSTNESS

Theorem (4.2) and Theorem (4.4) indicate that when $f_1$ is continuous a.e., the two feature spaces $(X_1, d_1)$ and $(X_2, d_2)$ or the functions $g_1$ and $g_2$ determine the strong-robustness of $f_1$. Based on Theorem (4.4), we can derive a corollary as follows (proof in Section 10.3.1):

**Corollary 5.1.** *When $f_1$ is continuous a.e., if $X_1 = \mathbb{R}^{n_1}$, $X_2 = \mathbb{R}^{n_2}$, $n_1 > n_2$, $X_2 \subsetneq X_1$, $d_1, d_2$ are norm functions, then $f_1(\cdot)$ is not strong-robust against adversarial examples.*

This corollary shows if unnecessary features (with regards to $X_2$) are selected in the feature selection step, then no matter how accurate the model is trained, it is not strong-robust to adversarial examples.

Figure 2 shows a situation that the oracle for the current task only needs to use one feature to classify samples correctly. A machine learning classifier extracts two features with one used by the oracle and the other is an extra unnecessary feature [11]. In $X_1$, $f_1$ (actually $c_1$) successfully classifies all the test inputs. However, it's very easy to find adversary examples satisfying Eq. (2.4) by only adding a small perturbation along the unnecessary feature dimension. In Figure 2, red circles show a few such adversarial examples. The adversarial examples are very close to seed samples in the oracle space. But they are predicted into a different class by $f_1$.

For many security sensitive applications, previous studies using state-of-art learning-based classifiers normally believe that adding more features is always helpful. Apparently, our corollary indicates that this thinking is wrong and can lead to their classifiers vulnerable to adversarial examples(Xu et al., 2016).

---

[11]Two features of $X_1$ actually positively correlate in Figure 2. However, the oracle does not need to use the second feature for making classification decision

As another example, multiple DNN studies about adversarial examples claim that adversarial examples are transferable among different DNN models. This can be explained by Figure 2 (when $X_1$ is a much higher-dimensional space). Since different DNN models learn over-complete feature spaces $\{X_1\}$, there is a high chance that these different $X_1$ involve a similar set of unnecessary features (e.g., the different learned features are correlated with others). Therefore the adversarial examples are generated along similar gradient directions. That is why many such samples can evade multiple DNN models.

### 5.2 Feature Space is More Important than Distance Measure using Norm Functions

Using Theorem (4.3), we obtain another corollary as follows (proof in Section 10.3.1):

**Corollary 5.2.** *When $f_1$ is continuous a.e., if $d_1$ and $d_2$ are norms and $X_1 = X_2 = \mathbb{R}^n$, then $f_1(\cdot)$ is strong-robust to adversarial examples.*

This corollary shows that if a learned classifier and its oracle share the same derived feature space ($X_1 = X_2$), the learned classifier is strong-robust when two metrics are both norm functions (even if not the same norm). We can call this corollary as "norm doesn't matter".

Many interesting phenomena can be answered by Corollary (5.2). For instance, for a norm regularized classifier, this corollary answers an important question that whether a different norm function will influence its robustness against adversarial examples. The corollary indicates that changing to a different norm function may not improve the robustness of the model under adversarial perturbation.

Summarizing Theorem (4.2), Theorem (4.4), Corollary (5.2) and Corollary (5.1), the robustness of a learned classifier is decided by two factors: (1) the difference between two derived feature spaces; and (2) the difference between the metric functions. Two corollaries show that the difference between the feature spaces is more important than the difference between the two metric functions.

### 5.3 Robustness and Generalization

In Table 3, we provide four situations in which the proposed theorems can be used to determine whether a classifier $f_1$ is strong-robust against adversarial examples or not.

- Case (I): If $f_1$ uses some unnecessary features, it will not be strong-robust to adversarial examples. It can not be an accurate predictor if $f_1$ misses some necessary features used by $f_2$.
- Case (II): If $f_1$ uses some unnecessary features, it will not be strong-robust to adversarial examples. It can be an accurate predictor if $f_1$ uses all the features used by $f_2$.
- Case (III): If $f_1$ and $f_2$ use the same set of features and nothing else, $f_1$ is strong-robust and can be accurate.
- Case (IV): If $f_1$ misses some necessary features and does not extract unnecessary features, $f_1$ is strong-robust (even tough its accuracy can not be good).

Table 3 provides a much better understanding of the relationship between robustness and accuracy. Two interesting cases from Table 3 are worth to be emphasized again: (1) If $f_1$ misses features used by $f_2$ and does not include unnecessary features (according to $X_2$), $f_1$ is strong-robust (even though it can not be accurate). (2) If $f_1$ extracts some extra unnecessary features, it will not be strong-robust (though it can be a very accurate predictor).

We want to emphasize that "$f_1$ is strong-robust" does not mean it is a good classifier. For example, a trivial example for strong-robust models is $f_1(x) \equiv 1, \forall x \in X$. However, it is a useless model since it doesn't have any prediction power. In an adversarial setting, we should aim to get a classifier that is both strong-robust and precise. A better feature learning function $g_1$ is exactly the solution that may achieve both goals.

Table 3 indicates that $c_1$ and $c_2$ do not influence the strong-robustness of $f_1$ when $f_1$ is continuous a.e. [12]. Figure 4 and Figure 5 further show two concrete example cases in which $f_1$ is strong-robust according to $f_2$. However, in both figures, $f_1$ is not accurate according to $f_2$.

---

[12] When $f_1$ is not continuous a.e., $c_1$ matters for the strong-robustness. See Section 13 for details.

Table 4: Connecting to relevant DNN hardening solutions. The experimental results of comparing different hardening solutions are shown in Figure 7, Figure 8, Table 10 and Table 11.

| | $x'$ | Loss $L_{f_1}(x, x')$ | On Layer |
|---|---|---|---|
| Stability training (Zheng et al., 2016) | random perturbation | $KL(f_1(x), f_1(x'))$ | Classification layer |
| Distributional training (Miyato et al., 2016) | adversarial perturbation | $KL(f_1(x), f_1(x'))$ | Classification layer |
| Adversarial training (Goodfellow et al., 2014) | adversarial perturbation | $L(f_1(x'), f_2(x))$ | Loss function |
| Large Adversarial training (Kurakin et al., 2016) | adversarial perturbation | $L(f_1(x'), f_2(x))$ | Loss function |
| Manifold based (Lee et al., 2015) | adversarial perturbation | $\| g_1(x) - g_1(x') \|_2$ | Layer before classification layer |
| Siamese Training (Section 12.3) | random perturbation | $\| g_1(x) - g_1(x') \|_2$ | Layer before classification layer |

# 6 Towards Principled Solutions for DNNs

Our theoretical analysis uncovers fundamental properties to explain adversarial examples. In this section, we apply them to analyze DNN classifiers. More specifically, (1) we find that DNNs are not strong-robust against adversarial examples; and (ii) we connect to possible hardening solutions and introduce principled understanding of these solutions.

## 6.1 Are State-of-the-Art DNNs Strong-Robust?

For DNN, it is difficult to derive a precise analytic form of $d_1$ (or $d'_1$). But we can observe some properties of $d_1$ through experimental results. Table 5, Table 6, Table 7 and Table 8 show properties of $d_1$ (and $d'_1$) resulting from performing testing experiments on four state-of-art DNN networks (details in Section 12.1). All four tables indicate that the accuracy of DNN models in the adversarial setting are quite bad. The performance on randomly perturbed inputs is much better than performance on maliciously perturbed adversarial examples.

The phenomenon we observed can be explained by Figure 3. Comparing the second column and the third column in four tables we can conclude that $d_1$ (and $d'_1$) in a random direction is larger than $d_1$ (and $d'_1$) in the adversarial direction. This indicates that a round sphere in $(X_1, d_1)$ (and $(X, d'_1)$) corresponds to a very thin high-dimensional ellipsoid in $(X, ||\cdot||)$ (illustrated by the left half of Figure 3). Figure 3 (I) shows a sphere in $(X, d'_1)$ and Figure 3 (III) shows a sphere in $(X_1, d_1)$. They correspond to the very thin high-dimensional ellipsoid in $(X, ||\cdot||)$ in Figure 3 (V). The norm function $||\cdot||$ is defined in space $X$ and is application-dependent. All four tables uses $||\cdot|| = ||\cdot||_\infty$.

Differently, for human oracles, a sphere in $(X, d'_2)$ (shown in Figure 3 (II)) or in $(X_2, d_2)$ (shown in Figure 3 (IV)) corresponds to an ellipsoid in $(X, ||\cdot||)$ not including very-thin-directions (shown in Figure 3 (VI)). When the attackers try to minimize the perturbation size using the approximated distance function $d_2 = ||\cdot||$, the thin direction of ellipsoid in Figure 3 (V) is exactly the adversarial direction.

## 6.2 Towards Principled Solutions

Our theorems suggest a list of possible solutions that may improve the robustness of DNN classifiers against adversarial samples. Options include such as:

**By learning a better $g_1$:** Methods like DNNs directly learn the feature extraction function $g_1$. Table 4 summarizes multiple hardening solutions (Zheng et al., 2016; Miyato et al., 2016; Lee et al., 2015) in the DNN literature. They mostly aim to learn a better $g_1$ by minimizing different loss functions $L_{f_1}(x, x')$ so that when $d_2(g_2(x), g_2(x')) < \epsilon$ (approximated by $(X, ||\cdot||)$), this loss $L_{f_1}(x, x')$ is small. Two major variations exist among related methods: the choice of $L_{f_1}(x, x')$ and the way to generate pairs of $(x, x')$. For instance, to reach the strong-robustness we can force to learn a $g_1$ that helps $(X, d_1')$ to be a finer topology than $(X_2, d_2')$. Section 12.3 explores this option ("Siamese training" in Table 4) through Siamese architecture. Experimentally Section 12.4 compares adversarial training, stability training and Siamese training on two state-of-the-art DNN image-classification tasks through performance against adversarial samples (details in Section 12.4). The hardening effects of these strategies vary from task to task, however, they all improve the base DNN models' performance in the adversarial setting.

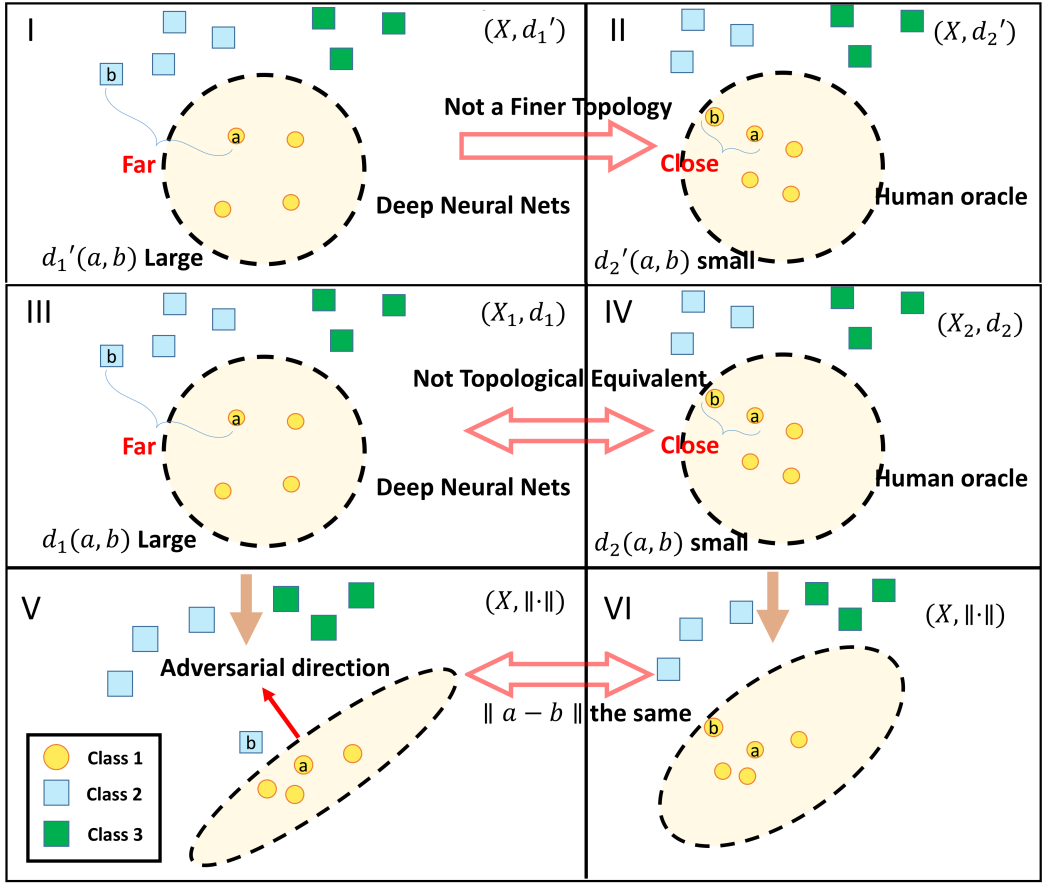

Figure 3: This figure shows one situation that $(X, d_1')$ is not a finer topology than $(X, d_2')$ (therefore, $(X_1, d_1)$ and $(X_2, d_2)$ are not topologically equivalent). According to Theorem (4.4), in this case, the DNN is vulnerable to adversarial examples. The two sample points $a$ and $b$ are close with regards to (w.r.t.) a norm $||\cdot||$ in $X$. They are also close w.r.t. $d_2$ in $(X_2, d_2)$ space and close w.r.t. $d_2'$ in $(X, d_2')$ space. But they are far from each other in the space of $(X, d_1')$ and in the space of $(X_1, d_1)$. In other words, while $d_2(a, b)$, $d_2'(a, b)$ and $||a - b||$ are small, $d_1(a, b)$ and $d_1'(a, b)$ are large. Clearly, DNN can be easily evaded by adding a small perturbation $||a - b||$ on sample $a$ or sample $b$. NOTE: it is normally difficult to get the analytic form of $(X_2, d_2)$ for most applications. Most previous studies (reviewed in Section 2.2) assume $(X_2, d_2)$ equals to $(X, ||\cdot||)$, where $||\cdot||$ is a norm function.

**By modifying unnecessary features:** As shown by Table 3, unnecessary features ruin the strong-robustness of learning-based classifiers. A simple way to remove the unrelated features is to identify which feature is unnecessary. In (Gao et al., 2017) the authors compare the difference between $g_1(x')$ and $g_1(x)$ from DNN. They hypothesize that those learned DNN feature dimensions (in $X_1$) changing rapidly are utilized by an adversary, and thus can be removed to improve the robustness of DNN model. Another efficient method is to substitute different values of features into several equivalent classes. By this way, the adversarial perturbation in the unnecessary feature dimensions can be squeezed into the same equivalent class. A recent study (Li & Vorobeychik, 2014) explored a similar strategy by using equivalent-feature-group to replace each word feature in a group, in order to improve the robustness of spam-email classifiers against evasion attacks.

### 6.2.1 CONNECTING TO PREVIOUS STUDIES HARDENING DNNs

Multiple hardening solutions (Zheng et al., 2016; Miyato et al., 2016; Lee et al., 2015) exist in the DNN literature. They mostly aim to learn a better $g_1$ by minimizing different loss functions $L_{f_1}(x, x')$ so that when $d_2(g_2(x), g_2(x')) < \epsilon$, this loss $L_{f_1}(x, x')$ is small. This might improve the the topological equivalence (or finer topology). Two major variations exist among related methods: the choice of $L_{f_1}(x, x')$ and the way to generate pairs of $(x, x')$.

- **Choice of loss function** $L_{f_1}(x, x')$: Siamese training (G) (Section 12.3) and (Lee et al., 2015) use $L_{f_1}(x, x') = d_1(g_1(x), g_1(x'))$. Siamese training (F) chooses $L_{f_1}(x, x') = dist(f_1(x), f_1(x'))$, where $dist(\cdot, \cdot)$ is a distance function measuring the difference between $f_1(x)$ and $f_1(x')$. If $f_1$ is continuous a.e., when $d_1(g_1(x), g_1(x'))$ is small → we get $dist(f_1(x), f_1(x'))$ is small. However, the reverse direction may not hold. Therefore, $L_{f_1}(x, x') = dist(f_1(x), f_1(x'))$ may not work for cases.

- **Generating pairs of** $(x, x')$: Another variation is the way of generating pairs of $(x, x')$ so that $d_2(g_2(x), g_2(x'))$ is small. There exist two common ways. One is generating $x'$ by adding a random (e.g. Gaussian) perturbation on $x$. The other one is generating the adversarial perturbation to get $x'$ from $x$.

Besides, (Zheng et al., 2016) uses $L_{f_1}(x, x') = KL(f_1(x), f_1(x'))$ and uses it as a regularization term adding onto the original training loss function. Its samples $x'$ are generated from original samples $x$ adding a small Gaussian noise. (Miyato et al., 2016) uses the similar loss function as (Zheng et al., 2016). But (Miyato et al., 2016) uses adversarial perturbed $x'$ from $x$. (Lee et al., 2015) uses $L_{f_1}(x, x') = d_1(g_1(x), g_1(x'))$ and $x'$s are generated $x$s by adding a small Gaussian noise. Recently proposed adversarial training (Goodfellow et al., 2014; Kurakin et al., 2016) uses $L_{f_1}(x, x') = L(f_1(x'), f_2(x))$ and uses adversarial perturbed $x'$ from $x$. These studies are summarized and compared in Table 4.

## 7 CONCLUSION

Adversarial examples are maliciously created inputs that lead a learning-based classifier to produce incorrect output labels. An adversarial example is often generated by adding small perturbations that appear unmodified to human observers. Recent studies that tried to analyze classifiers under adversarial examples are mostly empirical and provide little understanding of why. To fill the gap, we propose a theoretical framework for analyzing machine learning classifiers, especially deep neural networks (DNN) against such examples. This paper is divided into three parts. The first section provides a revised definition of adversarial examples by taking into account of the oracle of the task. The second section defines strong-robustness and provides the principled understanding of what makes a classifier strong-robust. The third section examines practical and theoretically grounded directions for understanding and hardening DNN models against adversarial examples. Future steps will include an empirical comparison to analyze recent literature using our theorems.

# 8 ASSUMPTION: ALMOST EVERYWHERE (A.E.) CONTINUITY

Most previous studies (Table 2) have made an important and implicit assumption about $f_1$ and $f_2$: $f_i$ is almost everywhere (a.e.) continuous. $i \in \{1, 2\}$.

**Definition 8.1.** *Suppose $f_i$ is the classification function. $f_i$ is continuous a.e., $i \in \{1, 2\}$, if $\forall x \in X$ a.e., $\exists \delta_i > 0$, such that $\forall x' \in X, d_i(g_i(x), g_i(x')) < \delta_i, f_i(x) = f_i(x')$.*

Illustrated in Figure 1, $d_i$ is the metric function (details in Section 4) $f_i$ uses to measure the similarity among samples in the space $X_i$. For notation simplicity, we use the term "continuous a.e." for "continuous almost everywhere"[13] in the rest of the paper. The above definition is a special case of almost everywhere continuity defined in (Folland, 2013) (see Definition (8.2) in Section 8.1), since we decompose $f_i$ in a certain way (see Figure 1). The a.e. continuity has a few indications, like:

$$\forall x, x' \in X, \quad \mathbb{P}(f_i(x) \neq f_i(x') | d_i(g_i(x), g_i(x')) < \delta_i) = 0 \tag{8.1}$$

See Section 8.1 for details of indications by a.e. continuity.

**$f_2$ is assumed continuous a.e. previously:** Most previous studies find "adversarial examples" by solving Eq. (2.1), instead of Eq. (2.2). This made an implicit assumption that if the adversarial example $x'$ is similar to the seed sample $x$, they belong to the same class according to $f_2$. This assumption essentially is: $f_2$ is almost everywhere (a.e.) continuous.

**$f_1$ is continuous a.e.:** Almost all popular machine learning classifiers satisfy the a.e. continuity assumption. For instance, a deep neural network is certainly continuous a.e.. Similarly to the results shown by (Szegedy et al., 2013), DNNs satisfy that $|f_1(x) - f_1(x')| \leq W \parallel x - x' \parallel_2$ where $W \leq \prod W_i$ and $W_i \geq ||(w_i, b_i)||_\infty$. Here $i = \{1, 2, \ldots, L\}$ representing $i$-th linear layer in NN. Therefore, $\forall \epsilon > 0$, let $\delta = \epsilon / W$. Then $|f_1(x) - f_1(x')| < \epsilon$ when $d_1(x, x') = \parallel x - x' \parallel_2 < \delta$. This shows that a deep neural network is almost everywhere continuous when $d_1(\cdot) = || \cdot ||_2$.

In Section 8.1, we show that if $f_1$ is not continuous a.e., it is not robust to any types of noise. Considering the generalization assumption of machine learning, machine learning classifiers should satisfy the continuity a.e. assumption. Section 8.2 provides two examples of how popular machine learning classifiers satisfy this assumption.

For the rare cases when $f_1$ is not continuous a.e., see next Section 13 discussing "boundary points" that matter for analyzing adversarial perturbations.

## 8.1 INDICATIONS FROM A.E. CONTINUITY ASSUMPTION

The a.e. continuity has a few indications,

- $X$ is not a finite space; and $\forall x, x' \in X, \mathbb{P}(f_i(x) = f_i(x') | d_i(g_i(x), g_i(x')) < \delta_i) = 1$
- It does not mean the function $f_1$ is continuous in every point in its feature space X;
- If a probability distribution admits a density, then the probability of every one-point set $\{a\}$ is zero; the same holds for finite and countable sets and the same conclusion holds for zero measure sets [14], for instance, straight lines or circle in $R^n$.
- The a.e. continuity follows the same property as density function: the probability of picking one-point set $\{x\}$ from the whole feature space is zero; the same holds for zero measure sets. This means: the probability of picking the discontinuous points (e.g., points on the decision boundary) is zero, because they are null sets.
- Most machine learning methods focus on $X = \mathbb{R}^p$ or space equivalent to $\mathbb{R}^p$ (e.g., $[0, 1]^p$) (see Appendix: Section 13.1). Most machine learning methods assume $f_1$ is continuous a.e. (see Appendix: Section 8.2).

**Definition 8.2.** *Suppose $(X, \mathcal{F}, \mathbb{P})$ is a probability space(for general definition, $(X, \Sigma, \mu)$ is a measure space), where $X$ is the sample space, a $\sigma$-algebra $\mathcal{F}$ is a collection of all the events and $\mathbb{P}$ is a probability measure defined in $X$ and $\mathcal{F}$. A property holds "almost everywhere" (a.e.) in $X$ if and only if the probability measure of the set for which the property holds equals 1.*

---

[13] The measure (e.g., Lebesgue measure) of discontinuous set is 0.

[14] Zero measure sets: also named as"Null set": https://en.wikipedia.org/wiki/Null_set

**Lemma 8.3.** *If the a.e. continuity assumption doesn't hold, there exists a non-zero measure set $\mathcal{D}$, such that*

$$\forall x \in \mathcal{D}, \exists x'$$
$$s.t. \ f_1(x) \neq f_1(x') \tag{8.2}$$
$$d_1(x, x') < \delta_1$$

*Proof.* Without it, for any test sample $x$, you can easily find a very similar sample $x'$ (i.e. for any small $\delta_1$, $d_1(x, x') < \delta_1$) such that $|f_1(x) - f_1(x')| > \epsilon$. In classification problems, this means that $f_1(x) \neq f_1(x')$(i.e. there exist very similar pair of two samples $x$ and $x'$ that have different labels for most $x \in X_1$). □

The Lemma (8.3) shows that $f_1$ is not robust to a random noise if we don't assume $f_1$ is continuous.

### 8.2 Most machine-learning classifiers satisfy the a.e. continuity assumption

Almost all popular machine learning classifiers satisfy the a.e. continuity assumption. For example,

- **Logistic regression for text categorization with a bag of word representation.**
  A classifier on a multivariate feature representation in which each feature representing (modified) counts of a word is naturally a.e. continuous. Since $\{x'|d_1(x, x') < \delta_1, x \neq x'\} = \emptyset$ when $\delta_1$ is small and $x, x'$ are mostly sparse vectors. Logistic regression with a bag of word representation is a continuous a.e. predictor.

- **Support Vector Machine with continuous feature representation.**
  Suppose we define $(X_1, d_1)$ by the $d_1^2(x, x') = k(x, x) + k(x', x') - 2k(x, x')$. Then support vector machine is a linear classifier on $(X_1, d_1)$. Thus, the SVM prediction function is continuous a.e. with $d_1$.

Most machine learning methods focus on the $\mathbb{R}^n$ space or the space equivalent to $\mathbb{R}^n$ (e.g., $[0, 1]^n$). For example, the sample space of image classification task intuitively is $255^p$, where $p$ is the number of features (e.g., $3 \times 224 \times 224$). However, people mostly rescale the raw image data samples into $X = [0, 1]^p$. Therefore, the sample space $X$ for $f_1$ for this case is $[0, 1]^p$.

## 9 More about Revised Formulation of Adversarial Examples

### 9.1 More about modeling oracle $f_2$

Though difficult, we want to argue that it is possible to theoretically model "oracles" for some state-of-the-art applications. For instance, as illustrated by the seminal cognitive neuroscience paper "untangling invariant object recognition" (DiCarlo & Cox, 2007) and its follow-up study (DiCarlo et al., 2012), the authors show that one can view the information processing of visual object recognition by human brains as the process of finding operations that progressively transform retinal representations into a new form of representation ($X_2$ in this paper), followed by the application of relatively simple decision functions (e.g., linear classifiers (Duda et al., 2012)). More specifically, in human and other primates, such visual recognition takes place along the ventral visual stream, and this stream is considered to be a progressive series of visual re-representations, from V1 to V2 to V4 to IT cortex (DiCarlo & Cox, 2007). Multiple relevant studies (e.g., (DiCarlo & Cox, 2007; Johnson, 1980; Hung et al., 2005)) have argued that this viewpoint of representation learning plus simple decision function is more productive than hypothesizing that brains directly learn very complex decision functions (highly non-linear) that operate on the retinal image representation. This is because the experimental evidence suggests that this view takes the problem apart in a way that is consistent with the architecture and response properties of the ventral visual stream. Besides, simple decision functions can be easily implemented in a single step of biologically plausible neuronal processing (i.e., a thresholded sum over weighted synapses).

As another example, the authors of (Xu et al., 2016) used genetic programming to find "adversarial examples" (by solving Eq. (2.2)) for a learning-based malicious-PDF classifier. This search needs an oracle to determine if a variant $x'$ preserves the malicious behavior of a seed PDF $x$ (i.e., $f_2(x) = f_2(x')$). The authors of (Xu et al., 2016) therefore used the Cuckoo sandbox (a malware analysis system through actual execution) to run a variant PDF sample in a virtual machine installed with a PDF reader and reported the behavior of the sample including network APIs calls. By comparing the behavioral signature of the original PDF malware and the manipulated variant, this oracle successfully determines if the malicious behavior is preserved from $x$ to $x'$. One may argue that "since Cuckoo sandbox works well for PDF-malware identification, why a machine-learning based detection system is even necessary?". This is because Cuckoo sandbox is computationally expensive and runs slow. For many security-sensitive applications about machines, oracles $f_2$ do exist, but machine-learning classifiers $f_1$ are used popularly due to speed or efficiency.

### 9.2 MORE ABOUT MODELING $f_1$: DECOMPOSITION OF $g_1$ AND $c_1$

It is difficult to decompose an arbitrary $f_1$ into $g_1 \circ c_1$. However, since in our context, $f_1$ is a machine learning classifier, we can enumerate many possible $g_1$ functions to cover classic machine learning classifiers.

- Various feature selection methods are potential $g_1$.
- For DNN, $g_1$ includes all the layers from input layer to the layer before the classification layer.
- In SVM, $X_1, d_1$ is decided by the chosen reproducing Hilbert kernel space.
- Regularization is another popular implicit feature extraction method. For example, $\ell_1$ regularization can automatically do the feature extraction by pushing some parameters to be 0.

## 10 APPENDIX: USING METRIC SPACE AND PSEUDO METRIC SPACES TO UNDERSTAND CLASSIFIERS' ROBUSTNESS AGAINST ADVERSARIAL EXAMPLES

### 10.1 METRIC SPACES AND TOPOLOGICAL EQUIVALENCE OF TWO METRIC SPACES

This subsection briefly introduces the concept of metric space and topological equivalence. A metric on a set/space $X$ is a function $d : X \times X \to [0, \infty]$ satisfying four properties: (1) non-negativity, (2) identity of indiscernibles, (3) symmetry and (4) triangle inequality. In machine learning, for example, the most widely used metric is Euclidean distance. Kernel based methods, such as SVM, kernel regression and Gaussian process, consider samples in a Reproducing kernel Hilbert space (RKHS). The metric in a RKHS is naturally defined as: $d^2(x, y) = K(x, x) + K(y, y) - 2K(x, y)$, in which $K(\cdot, \cdot)$ is a kernel function.

Now we present an important definition, namely that of "topological equivalence", that can represent a special relationship between two metric spaces.

**Definition 10.1.** ***Topological Equivalence** (Kelley, 1975)*

*A function or mapping $h(\cdot)$ from one topological space to another is continuous if the inverse image of any open set is open. If this continuous function is one-to-one and onto, and the inverse of the function is also continuous, then the function is called a homeomorphism and the domain of the function, in our case $(X_1, d_1)$, is said to be homeomorphic to the output range, e.g., here $(X_2, d_2)$. In other words, metric space $(X_1, d_1)$ is topologically equivalent to the metric space $(X_2, d_2)$.*

We can state this definition as the following equation:

$$\exists h : X_1 \to X_2, \forall x_1, x_1' \in X_1,$$
$$h(x_1) = x_2, h(x_1') = x_2' \qquad (10.1)$$
$$d_1(x_1, x_1') < \delta_1 \Leftrightarrow d_2(x_2, x_2') < \delta_2$$

Here $h$ is continuous, one-to-one and onto. $\delta_1$ and $\delta_2$ are two small constants.

### 10.2 PSEUDOMETRIC SPACES AND FINER TOPOLOGY AMONG PSEUDOMETRIC SPACES

We have briefly reviewed the concept of metric space in Section 10.1 and proposed the related Theorem (4.2) in Section 4.3. This is partly because the concept of metric space has been widely used in many machine learning models, such as metric learning (Xing et al., 2003). Theorem (4.2) and related analysis indicate that feature spaces $X_1$ and $X_2$ (See Figure 1) are key determining factors for deciding learning model's strong-robustness.

However, it is difficult to get the analytic form of $X_2$ in most applications (e.g., when an oracle $f_2$ is a human annotator). In fact, most previous studies (reviewed in Section 2.2) assume $(X_2, d_2)$ equals to $(X, ||\cdot||)$, where $||\cdot||$ is a norm function. Therefore, we want to extend our analysis and results from the implicit feature space $X_2$ to the original feature space $X$.

When we extend the analysis to the original space $X$, it is important to point out that the distance function measuring sample similarity for a learned predictor $f_1$ in the original space $X$ may not be a metric. The distance function in the original feature space $X$ for oracle $f_2$ may not be a metric as well. This is because the distance between two different samples in the original space $X$ may equal to 0. Because two different samples may be projected into the same point in $X_1$ or $X_2$. For example, a change in one pixel of background in an image does not affect the prediction of $f_1$ or $f_2$ since the $g_1$ and $g_2$ have already eliminated that (irrelevant) feature. This property contradicts the identity of indiscernibles assumption for a metric function. Therefore we need a more general concept of the distance function for performing theoretical analysis in the original space $X$. By using the concept of **Pseudometric Space**[15], we derive another important theorem about strong-robustness.

**Pseudometric:** If a distance function $d' : X \times X \to [0, \infty]$ has the following three properties: (1) non-negativity, (2) symmetry and (3) triangle inequality, we call $d$ is a pseudometric or generalized metric. The space $(X, d')$ is a pseudometric space or generalized metric space. It is worth to point out that the generalized metric space is a special case of topological space and metric space is a special case of pseudometric space.

**Why Pseudometric Space:** As shown in Figure 1, we can decompose a common machine learning classifier $f_1 = c_1 \circ g_1$, where $g_1 : X \to X_1$ represents the feature extraction and $c_1 : X_1 \to Y$ performs the operation of classification. Assume there exists a pseudometric $d'_1(\cdot, \cdot)$ on $X$ and a metric $d_1(\cdot, \cdot)$ defined on $X_1$[16], so that $\forall x, x' \in X$,

$$d'_1(x, x') = d_1(g_1(x), g_1(x')). \tag{10.2}$$

Since $d_1$ is a metric in $X_1$, $d'_1$ fulfills the (1) non-negativity, (2) symmetry and (3) triangle inequality properties. However, $d'_1$ may not satisfy the identity of indiscernible property (i.e., making it not a metric). For example, suppose $g_1$ only selects the first three features from $X$. Two samples $x$ and $x'$ have the same value in the first three features but different values in the rest features. Clearly, $x \neq x'$, but $d'_1(x, x') = d_1(g_1(x), g_1(x')) = 0$. This shows that $d'_1(\cdot, \cdot)$ is a pseudometric but not a metric in $X$. Similarly, a pseudometric $d'_2$ for the oracle can be defined as follow:

$$d'_2(x, x') = d_2(g_2(x), g_2(x')). \tag{10.3}$$

To analyze the strong robustness problem in the original feature space $X$, we assume it to be a generalized metric (pseudometric) space $(X, d'_1)$ for $f_1$ and a generalized metric (pseudometric) space $(X, d'_2)$ for $f_2$. Now we can analyze $f_1$ and $f_2$ on the same feature space $X$ but relate to two different pseudometrics. This makes it possible to define a sufficient and necessary condition for determining the strong robustness of $f_1$ against adversarial perturbation.

Before introducing this condition, we need to briefly introduce the definition of topology and finer/coarser topology here:

**Definition 10.2.** *A topology $\tau$ is a collection of open sets in a space $X$.*

A topology $\tau$ is generated by a collection of open balls $\{B(x, \delta_1)\}$ where $x \in X$ and $B(x, \delta_1) = \{z | d(x, z) < \delta_1\}$. The collection contains $\{B(x, \delta_1)\}$, the infinite/finite number of the union of balls, and the finite number of intersection of them.

**Definition 10.3.** *Suppose $\tau_1$ and $\tau_2$ are two topologies in space $X$. If $\tau_2 \subseteq \tau_1$, the topology $\tau_2$ is called a coarser (weaker or smaller) topology than the topology $\tau_1$, and $\tau_1$ is called a finer (stronger or larger) topology than $\tau_2$.*

---

[15]The crucial problem of the original sample space $X$ is that it's difficult to strictly define a metric on the original feature space.

[16]$d_1(\cdot, \cdot)$ on $X_1$ satisfies four properties:(1) non-negativity, (2) identity of indiscernibles, (3) symmetry and (4) triangle inequality.

### 10.3 Proofs for Theorems and Corollaries

In this section, we provide the proofs for Theorem (4.2), Corollary (5.2), Theorem (4.4), and Corollary (5.1). We first prove Theorem (4.4) and Corollary (5.1). Since "topological equivalence" is a stronger condition than "finer topology", Theorem (4.2) and Corollary (5.2) are straightforward.

#### 10.3.1 Proof of Theorem (4.4) when $f_2$ is continuous a.e.

*Proof.* Let $S_1 = \{B_1(x, \epsilon)\}$ and $S_2 = \{B_2(x, \epsilon)\}$, where $B_1(x, \epsilon) = \{y | d'_1(x, y) < \epsilon\}$ and $B_2(x, \epsilon) = \{y | d'_2(x, y) < \epsilon\}$. Then $S_1 \subset \tau_1$ and $S_2 \subset \tau_2$. In fact, $\tau_1$ and $\tau_2$ are generated by $S_1$ and $S_2$. $S_1$ and $S_2$ are bases of $(X, \tau_1)$ and $(X, \tau_2)$.

- First, we want to prove that given $\delta_2 > 0$, $\exists \delta_1 > 0$ such that if $d'_2(x, x') \leq \delta_2$, then $d'_1(x, x') \leq \delta_1$. Consider a pair of samples $(x, x')$ and $d'_2(x, x') \leq \delta_2$. $x, x' \in B_2(x, \delta_2)$. Of course, $B_2(x, \delta_2) \in \tau_2$. Suppose the $(X, d'_1)$ is a finer topology than $(X, d'_2)$. Then $B_2(x, \delta_2) \in \tau_1$. You can find $B_1(x_0, \delta_1/2) \in \tau_1$ such that $\bar{B}_2(x, \delta_2) \subset \bar{B}_1(x_0, \delta_1/2)$, where $\bar{B}_2(x, \delta_2)$ is the closure of $B_2(x, \delta_2)$. Therefore $d'_1(x, x') \leq \delta_1$.
  Based on a.e. continuity assumption of $f_1$, since $d'_1(x, x') \leq \delta$, $f_1(x) = f_1(x')$ a.e. . This means that $\mathbb{P}(f_1(x) = f_1(x') | d_2(g_2(x), g_2(x')) < \delta_2) = 1$, which is our definition of strong-robustness.
- Next, we want to show that if $f_1$ is strong-robust, then $\tau_1$ is a finer topology than $\tau_2$.
  Suppose $f_1$ is strong-robust, we need to prove that $\forall \delta_2 > 0$, $\exists \delta_1 > 0$ such that if $d'_2(x, x') \leq \delta_2$, then $d'_1(x, x') \leq \delta_1$.
  Assume $\tau_1$ is not a finer topology than $\tau_2$. This means there exists a $B_2(x, \delta_2)$ such that $B_2(x, \delta_2) \notin \tau_1$. Therefore $\forall \delta_1 > 0$, there exists $x' \in B_2(x, \delta_2)$ such that $d'_2(x, x') < \delta_2$ and $d'_1(x, x') > \delta_1$. Based on a.e. continuity assumption of $f_1$, $d'_1(x, x') > \delta_1$ indicates that $f_1(x) \neq f_1(x')$. This contradicts the strong-robust assumption. Thus, $\tau_1$ is a finer topology than $\tau_2$.

□

**Proof of Theorem (4.4) when $f_2$ is not continuous a.e.**

*Proof.* Let $S_1 = \{B_1(x, \epsilon)\}$ and $S_2 = \{B_2(x, \epsilon)\}$, where $B_1(x, \epsilon) = \{y | d'_1(x, y) < \epsilon\}$ and $B_2(x, \epsilon) = \{y | d'_2(x, y) < \epsilon\}$. Then $S_1 \subset \tau_1$ and $S_2 \subset \tau_2$. In fact, $\tau_1$ and $\tau_2$ are generated by $S_1$ and $S_2$. $S_1$ and $S_2$ are bases of $(X, \tau_1)$ and $(X, \tau_2)$.

- First, we want to prove that given $\delta_2 > 0$, $\exists \delta_1 > 0$ such that if $d'_2(x, x') \leq \delta_2$, then $d'_1(x, x') \leq \delta_1$. Consider a pair of samples $(x, x')$ and $d'_2(x, x') \leq \delta_2$. $x, x' \in B_2(x, \delta_2)$. Of course, $B_2(x, \delta_2) \in \tau_2$. Suppose the $(X, d'_1)$ is a finer topology than $(X, d'_2)$. Then $B_2(x, \delta_2) \in \tau_1$. You can find $B_1(x_0, \delta_1/2) \in \tau_1$ such that $\bar{B}_2(x, \delta_2) \subset \bar{B}_1(x_0, \delta_1/2)$. Therefore $d'_1(x, x') \leq \delta_1$.
  Based on a.e. continuity assumption of $f_1$, since $d'_1(x, x') \leq \delta_1$, $f_1(x) = f_1(x')$ a.e. . This means that $\mathbb{P}(f_1(x) = f_1(x') | f_2(x) = f_2(x'), d_2(g_2(x), g_2(x')) < \delta_2) = 1$, which is our definition of strong-robustness.
- Next, we want to show that if $f_1$ is strong-robust, then $\tau_1$ is a finer topology than $\tau_2$.
  Suppose $f_1$ is strong-robust, we need to prove that $\forall \delta_2 > 0$, $\exists \delta_1 > 0$ such that if $d'_2(x, x') \leq \delta_2$, then $d'_1(x, x') \leq \delta_1$.
  Assume $\tau_1$ is not a finer topology than $\tau_2$. This means there exists a $B_2(x, \delta_2)$ such that $B_2(x, \delta_2) \notin \tau_1$. Therefore $\forall \delta_1 > 0$, there exists $x' \in B_2(x, \delta_2)$ such that $d'_2(x, x') < \delta_2$ and $d'_1(x, x') > \delta_1$. Based on a.e. continuity assumption of $f_1$, $d'_1(x, x') > \delta_1$ indicates that $f_1(x) \neq f_1(x')$. This contradicts the strong-robust assumption. Thus, $\tau_1$ is a finer topology than $\tau_2$.

□

### 10.3.2  Proof of Theorem (4.5)

*Proof.* In Section 10.3.1, we have already proved that if the $(X, d_1')$ is a finer topology than $(X, d_2')$, then we can have that $\forall$ pair $(x, x')$ $(x, x' \in X)$ $d_2'(x, x') \le \delta_2$, then $d_1'(x, x') \le \delta_1$. Therefore,

$$\mathbb{P}(f_1(x) = f_1(x')|f_2(x) = f_2(x'), d_2'(x, x') < \delta_2)$$
$$=1 - \mathbb{P}(f_1(x) \ne f_1(x')|f_2(x) = f_2(x'), d_2'(x, x') < \delta_2)$$
$$=1 - \mathbb{P}(f_1(x) \ne f_1(x')|f_2(x) = f_2(x'), d_1(x, x') < \delta_1, \tag{10.4}$$
$$d_2'(x, x') < \delta_2)$$
$$>1 - \eta$$

$\square$

### 10.3.3  Proof of Theorem (4.3)

*Proof.* Since $(X_1, d_1)$ and $(X_2, d_2)$ are topologically equivalent. $\mathbb{P}(f_1(x) \ne f_1(x')|f_2(x) = f_2(x'), d_1(g_1(x), g_1(x')) < \delta_1) = \mathbb{P}(f_1(x) \ne f_1(x')|f_2(x) = f_2(x'), d_2(g_2(x), g_2(x')) < \delta_2)$. Therefore,

$$\mathbb{P}(f_1(x) = f_1(x')|f_2(x) = f_2(x'), d_2(g_2(x), g_2(x')) < \delta_2)$$
$$=1 - \mathbb{P}(f_1(x) \ne f_1(x')|f_2(x) = f_2(x'),$$
$$d_2(g_2(x), g_2(x')) < \delta_2)$$
$$=1 - \mathbb{P}(f_1(x) \ne f_1(x')|f_2(x) = f_2(x'), \tag{10.5}$$
$$d_1(g_1(x), g_1(x')) < \delta_1, d_2(g_2(x), g_2(x')) < \delta_2)$$
$$>1 - \eta$$

$\square$

### 10.3.4  Proof of Theorem (4.2)

*Proof.* Since $f_1$ is continuous a.e., $\mathbb{P}(f_1(x) = f_1(x')|f_2(x) = f_2(x'), d_1(g_1(x), g_1(x') < \delta_1, d_2(g_2(x), g_2(x')) < \delta_2) = 0$. Therefore, by Section 10.3.3, $\mathbb{P}(f_1(x) = f_1(x')|f_2(x) = f_2(x'), d_2(g_2(x), g_2(x')) < \delta_2) = 1$. $\square$

### 10.3.5  Proof of Corollary (5.2)

*Proof.* By (Kelley, 1975), we know that if $d_1$ and $d_2$ are norms in $\mathbb{R}^n$, $(\mathbb{R}^n, d_1)$ and $(\mathbb{R}^n, d_2)$ are topological equivalent. Therefore, we have the conclusion. $\square$

### 10.3.6  Proof of Corollary (5.1)

*Proof.* Suppose $n_1 > n_2$ and $X_2 \subset X_1$. $(X, d_2')$ is a finer topology than $(X, d_1')$. Therefore $(X, d_1')$ is not a finer topology than $(X, d_2')$, which indicates that $f_1$ is not strong-robust against adversarial examples. $\square$

## 11  More about Principled Understanding from Proposed Theorems

### 11.1  More Examples about $g_1$ Matters for Strong-robustness and $c_1$ Not

Figure 4 uses an example to illustrate Table 3 Case (III) when $f_1$ is strong-robust. We show one case of $X_1 = X_2 = \mathbb{R}^2$ and $f_1$, $f_2$ are continuous a.e.. In terms of classification, $f_1$ (green boundary line) is not accurate according to $f_2$ (red boundary line).

Figure 5 uses an example figure to illustrate Table 3 Case (IV) when $f_1$ is strong-robust. We show one case of $1 = n_1 < n_2 = 2$, $X_1 \subset X_2$ and $f_1$, $f_2$ are continuous a.e.. In terms of classification, $f_1$ (green boundary line) is not accurate according to $f_2$ (red boundary line).

All pairs of test samples $(x, x')$ can be categorized into the three cases shown in both figures.

- Test-case (a) is when $x$ and $x'$ are predicted as the same class by both. $f_1$ gets correct predictions according to $f_2$. There exist no adversarial examples.
- Test-case (b) is when $x$ and $x'$ are predicted as the same class by both. But $f_1$ gets incorrect predictions according to $f_2$. There exist no adversarial examples.
- Test-case (c) shows when $f_1(x) \neq f_1(x')$, $d_2(x, x') < \delta_2$ and $f_2(x) = f_2(x')$. This case is explained in Section 13. Essentially, this is about "Boundary based adversarial examples" and can only attack points whose distance to the boundary of $f_1$ is smaller than $\delta_2$ ($f_1(x) \neq f_1(x')$, $d_2(x, x') < \delta_2$ and $f_2(x) = f_2(x')$). When $f_1$ is continuous a.e., the probability of this set is 0.

Clearly from the two figures, $c_1$ does not determine the strong-robustness of $f_1$.

## 11.2 More about Extra Unnecessary Features Ruin Strong-robustness

In real-world applications, such attacks can be, for example, adding words with a very tiny font size in a spam E-mail, that is invisible to a human annotator. When a learning-based classifier tries to utilize such extra words (unnecessary for human), it can lead to many easily generated adversarial emails.

As another example, one previous study (Xu et al., 2016) shows that a genetic-programming based adversarial example strategy can always evade two state-of-art learning-based PDF-malware classifiers (with "100%" evasion rates). The reason behind such good evasion rates is the Condition (5.1). Both state-of-art PDF-malware classifiers have used many superficial features (e.g., a feature representing "is there a long comment section") that are not relevant to "the malicious property" of a PDF sample at all !

## 11.3 When $f_1$ Continuous a.e., Either Strong-robust or Not robust at all a.e.

Table 3 indicates that training a strong-robust and accurate classifier in practice is extremely difficult. For instance, Figure 2 shows only one extra irrelevant feature, which does not hurt accuracy, makes the classifier not robust to adversarial perturbation at all (i.e., for samples a.e. in $X$, easy to find its adversarial examples.).

When $f_1$ is continuous a.e., $\mathbb{P}(f_1(x) = f_1(x') | f_2(x) = f_2(x'), d_2(g_2(x), g_2(x')) < \delta_2)$ equals to either 1 or 0. This means $f_1$ is either strong-robust or not robust under AN at all a.e.. One case with this probability as 0 is illustrated by Figure 2. Case (III) and Case (IV) from Table 3 have this probability equaling to 1.

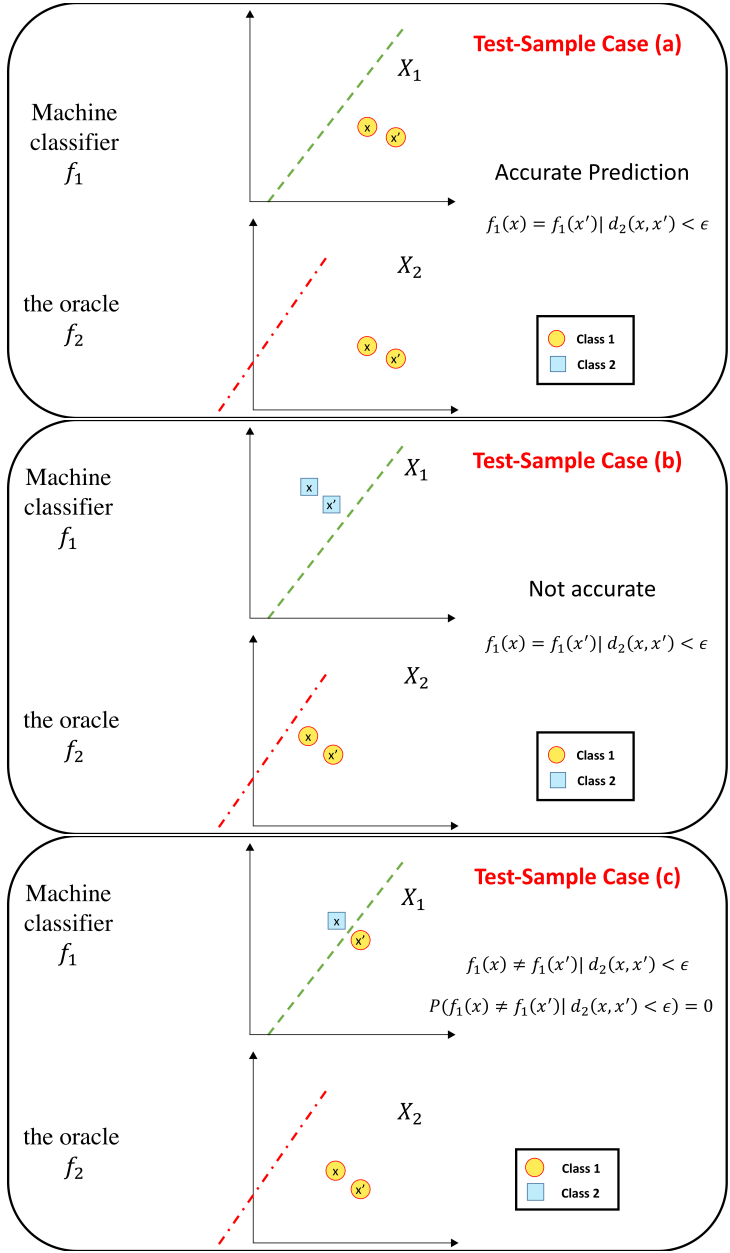

Figure 4: An example figure illustrating Table 3 Case (III) when $f_1$ is strong-robust. We assume $c_1$ and $c_2$ as linear classification functions. We show one case of $X_1 = X_2 = \mathbb{R}^2$ and $f_1$, $f_2$ are continuous a.e.. In terms of classification, $f_1$ (green boundary line) is not accurate according to $f_2$ (red boundary line). All pairs of test samples $(x, x')$ can be categorized into the three cases shown in this figure. Test-case (a): $f_1$ and $f_2$ assign the same classification label (yellow circle) on $x$ and $x'$. $x$ and $x'$ are predicted as the same class by both. Test-case (b): $f_1$ assigns the class of "blue square" on both $x$ and $x'$. $f_2$ assigns the class of "yellow circle" on both $x$ and $x'$. Test-case (c): $f_2$ assigns the class of "yellow circle" on both $x$ and $x'$. However, $f_1$ assigns the class of "blue square" on $x$ and assigns a different class of "yellow circle" on $x'$. This case has been explained in Section 13.

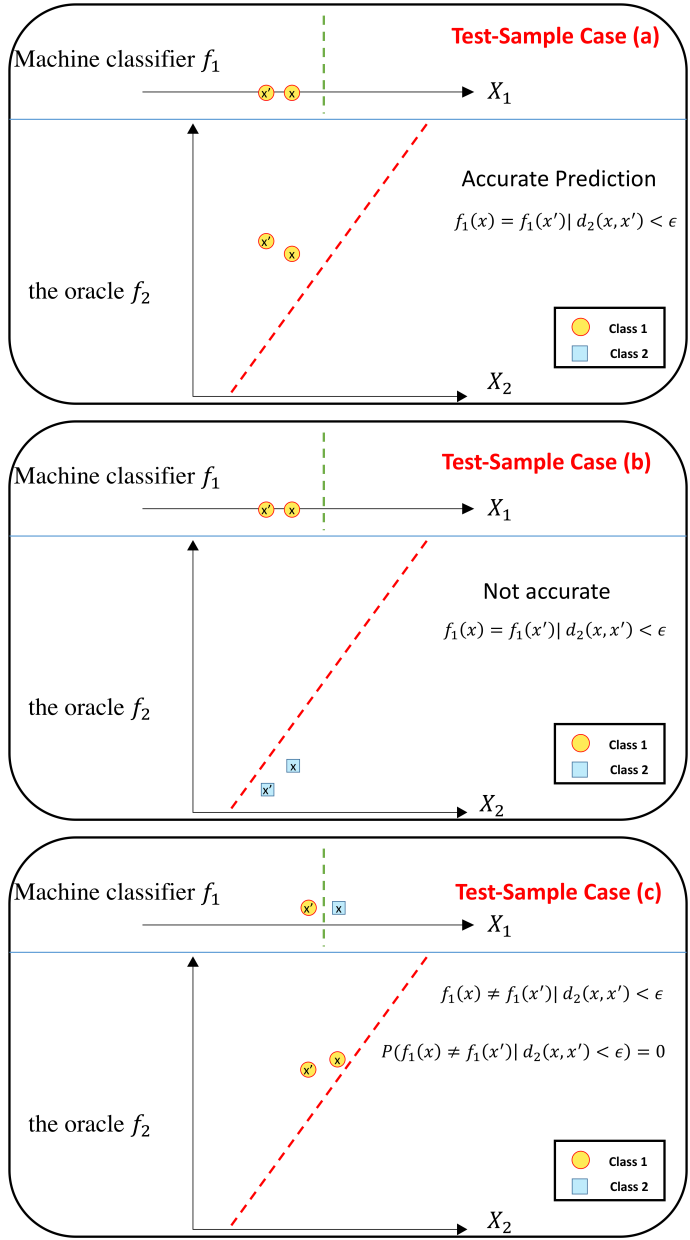

Figure 5: An example figure illustrating Table 3 Case (IV) when $f_1$ is strong-robust. We assume $c_1$ and $c_2$ as linear classification functions. We show one case of $1 = n_1 < n_2 = 2$, $X_1 \subset X_2$ and $f_1$, $f_2$ are continuous a.e.. In terms of classification, $f_1$ (green boundary line) is not accurate according to $f_2$ (red boundary line). All pairs of test samples $(x, x')$ can be categorized into the three cases shown in this figure. Test-case (a): $f_1$ and $f_2$ assign the same classification label (yellow circle) on $x$ and $x'$. $x$ and $x'$ are predicted as the same class by both. Test-case (b): $f_1$ assigns the class of "yellow circle" on both $x$ and $x'$. $f_2$ assigns the class of "blue square" on both $x$ and $x'$. Test-case (c): $f_2$ assigns the class of "yellow circle" on both $x$ and $x'$. However, $f_1$ assigns the class of "blue square" on $x$ and assigns a different class of "yellow circle" on $x'$. This case can be explained in Section 13.

## 12    MORE ABOUT TOWARDS PRINCIPLED SOLUTIONS FOR DNNS

Researchers have proposed different strategies to generate adversarial examples attacking deep neural networks (e.g., (Szegedy et al., 2013; Nguyen et al., 2015; He et al., 2015; Papernot et al., 2016a; Moosavi-Dezfooli et al., 2015; Papernot et al., 2015b)). Previous studies mostly focus on an image classification, therefore for these domains our symbols mean:

- $f_1(\cdot)$: $f_1(\cdot)$ is a DNN classifier with multiple layers, including linear perceptron layers, activation layers, convolutional layers and softmax decision layer.
- $(X_1, d_1)$: $X_1$ denotes the feature space discovered by the layer right before the last fully connected layer. This feature space is automatically extracted from the original image space (e.g., RGB representation) by the DNN. $(X, d'_1)$ is defined by $d_1$ using Eq. (10.2).
- $(X_2, d_2)$: $X_2$ denotes the feature space that oracle (e.g., human annotators) used to decide ground-truth labels of training images. For example, a human annotator needs to recognize a hand-written digit "0". $X_2$ includes what patterns he/she needs for such a decision. $(X, d'_2)$ is defined by $d_2$ using Eq. (10.3)

### 12.1    MORE ABOUT ARE STATE-OF-THE-ART DEEP NEURAL NETS STRONG-ROBUST ?

We can observe some properties of $d_1$ through experimental results. Table 5,Table 6,Table 7 and Table 8 show properties of $d_1$ (and $d'_1$) resulting from performing testing experiments on four state-of-art DNN networks.

In Table 9, the model we use is a 200-layer residual network (He et al., 2015) trained on Imagenet dataset (Deng et al., 2009) by Facebook[17]. We generate two types of test samples from 50000 images in the validation set of Imagenet: (1) 50000 randomly perturbed images. The random perturbations on each image are generated by first fixing the perturbation value on every dimension to be the same, and then randomly assigning the sign on every dimension as $+$ or $-$ (with probability $1/2$). In this way, the size of the perturbation can be described by $||x - x'||_\infty$ that we name as the level of **attacking power** ( later defined in Eq. (12.6)). (2) 50000 adversarially perturbed images. We use the fast-gradient sign method (introduced in Section 3.2) to generate such adversarial perturbations on each seed image. The "attacking power" of such adversarial perturbations uses the same formula as Eq. (12.6). The first column of Table 9 shows different attack powers (Eq. (12.6)) we use in the experiment. The second column shows the accuracy of running the DNN model on the first group of image samples and the third column shows the accuracy of running the DNN model on the second group of image samples.

Table 6,Table 7 and Table 8 repeat similar experiments on three other DNN models: overfeat network(Sermanet et al., 2013), the residual network(He et al., 2015) and the VGG model (Simonyan & Zisserman, 2014). The conclusion is consistent across all four models.

### 12.2    A NOVEL EVALUATION METRIC "ADVERSARIAL ROBUSTNESS OF CLASSIFIERS (ARC)" TO QUANTIFY THE MODEL ROBUSTNESS AGAINST ADVERSARIAL EXAMPLES

Our theoretical analysis indicates that strong-robustness is a strong condition of machine learning classifiers and requires thorough understanding of oracle. Since many state-of-the-art learning models, including many DNNs, are not strong-robust, it is important to understand and quantify how far they are away from strong-robustness.

This section proposes a new evaluation measure "Adversarial Robustness of Classifiers (ARC)" to quantify how far a classifier is away from the strong-robustness. This quantitative measure considers both the predictor $f_1$ and the oracle $f_2$. By design, a classifier ($f_1$)'s ARC achieves the maximum (1 since ARC is rescaled to $[0, 1]$) if and only if $f_1$ is strong-robust (see Theorem (12.3)).

#### 12.2.1    DEFINE ARC AND ARCA

We name such situations as "weak-robustness" and propose a quantitative measure to describe how robust a classification model is against adversarial examples. The proposed measure "Adversarial Robustness of Classifiers (ARC)" considers both the predictor $f_1$ and the oracle $f_2$ (introduced in

---

[17]https://github.com/facebook/fb.resnet.torch

Table 5: Accuracy of the deep residual network(He et al., 2015) obtained from two noise-perturbed testing cases. The second column shows the result on randomly perturbed samples, and the third column shows the result on adversarially perturbed samples.

| Attack power (defined in Eq. (12.6)) | Test accuracy on randomly perturbed samples | Test accuracy on adversarially perturbed samples |
|---|---|---|
| 0 | 0.9411 | 0.9411 |
| 1 | 0.9409 | **0.5833** |
| 5 | 0.9369 | **0.3943** |
| 10 | 0.9288 | **0.3853** |

Table 6: Accuracy of the overfeat network(Sermanet et al., 2013) obtained from two noise-perturbed testing cases. The second column shows the result on randomly perturbed samples, and the third column shows the result on adversarially perturbed samples.

| Attack power (defined in Eq. (12.6)) | Test accuracy on randomly perturbed samples | Test accuracy on adversarially perturbed samples |
|---|---|---|
| 0 | 0.7944 | 0.7944 |
| 1 | 0.7923 | **0.5922** |
| 5 | 0.7844 | **0.4270** |
| 10 | 0.7762 | **0.3485** |

Section 2.2). By design, a classifier ($f_1$)'s ARC achieves the maximum (1 since ARC is rescaled to $[0, 1]$) if and only if $f_1$ is strong-robust against adversarial examples and is based on the expectation of how difficult it is to generate adversarial examples.

**Definition 12.1.** *Adversarial Robustness of Classifiers (ARC)*

*By adding the constraint $d_2(x, x') < \delta_2$ into Eq. (2.2) (our general definition of adversarial examples) and taking the expectation of $d_2$ between adversarial example and seed sample, we define a measure quantifying the robustness of machine learning classifiers against adversarial examples.*

$$ARC(f_1, f_2) = \mathbb{E}_{x \in X}[d_2(x, x')]$$
$$x' = \underset{t \in X}{\operatorname{argmin}} \, d_2(x, t)$$
$$\textit{Subject to: } f_1(x) \neq f_1(t)$$
$$d_2(x, t) < \delta_2$$

(12.1)

Here for the case that $x'$ doesn't exist, we assume $d_2(x, x') = \delta_2$.

Two recent studies (Moosavi-Dezfooli et al., 2015; Papernot et al., 2015b) propose two similar measures both assuming $d_2$ as norm functions, but do not consider the importance of an oracle. More importantly, (Papernot et al., 2015b) does not provide any computable way to calculate the measure. In (Moosavi-Dezfooli et al., 2015), the measure is normalized by the size of the test samples, while no evidence exists to show that the size of perturbation is related to the size of test samples.

The fact that previous measures neglect the oracle $f_2$ leads to a severe problem: the generated adversarial examples are not necessarily valid. This is because if the size of perturbation is too large, oracle $f_2$ may classify the perturbed sample into a different class (different from the class of the seed sample).

This motivates us to design a computable criteria to estimate Definition (12.1). For instance, for image classification tasks, we can choose $d_2 = || \cdot ||_\infty$ as an example. Then in Eq. (12.1), to estimate of $\mathbb{E}[||x - x'||_\infty]$, we need to make some assumptions. Assume that there exists a threshold $\delta_2$, that any perturbation larger than $\delta_2$ will change the classification of oracle $f_2$. That is if $||x - x'||_\infty \geq \delta_2$, then $f_2(x) \neq f_2(x')$. More concretely, for image classification tasks, as the input space is discrete (with every dimension ranging from 0 to 255), ARC can be estimated by the following Eq. (12.2):

$$
\begin{aligned}
ARC_\infty(f_1, f_2) = & \mathbb{E}[\| x - x' \|_\infty] = \sum_{i=1}^{\delta_2 - 1} i \mathbb{P}(\| x - x' \|_\infty = i) \\
& + \delta_2 \mathbb{P}(f_1(x) = f_1(t), \forall \| x - t \|_\infty < \delta_2). \\
& x' = \operatorname*{argmin}_{t \in X} d_2(x, t) \\
\text{Subject to: } & f_1(x) \neq f_1(t) \\
& f_2(x) = f_2(t)
\end{aligned}
\tag{12.2}
$$

Table 7: Accuracy of the residual network(He et al., 2015) obtained from two noise-perturbed testing cases in CIFAR-10 dataset (Krizhevsky & Hinton, 2009). The second column shows the result on randomly perturbed samples, and the third column shows the result on adversarially perturbed samples.

| Attack power (defined in Eq. (12.6)) | Test accuracy on randomly perturbed samples | Test accuracy on adversarially perturbed samples |
|---|---|---|
| 0 | 0.9431 | 0.9431 |
| 1 | 0.9431 | **0.9294** |
| 5 | 0.9429 | **0.6815** |
| 10 | 0.943 | **0.2961** |

Table 8: Accuracy of the wide residual network(Zagoruyko & Komodakis, 2016) obtained from two noise-perturbed testing cases in CIFAR-10 dataset (Krizhevsky & Hinton, 2009). The second column shows the result on randomly perturbed samples, and the third column shows the result on adversarially perturbed samples.

| Attack power (defined in Eq. (12.6)) | Test accuracy on randomly perturbed samples | Test accuracy on adversarially perturbed samples |
|---|---|---|
| 0 | 0.953 | 0.953 |
| 1 | 0.953 | **0.8527** |
| 5 | 0.953 | **0.4718** |
| 10 | 0.953 | **0.2529** |

**Definition 12.2.** *Adversarial Robustness of Classifiers with Accuracy (ARCA)*

*As we have discussed in the Section 5, both accuracy and robustness are important properties in determining whether a classification model is preferred or not. Therefore we combine accuracy and ARC into the following unified measure ARCA:*

$$
ARCA(f_1) = Accuracy(f_1) \times \frac{ARC(f_1, f_2)}{\delta_2}
\tag{12.3}
$$

### 12.2.2   ARC AND STRONG-ROBUSTNESS

It is important to understand the relationship between strong-robustness and weak-robustness. We provide an important theorem as follows that clearly shows the weak-robustness is quantitatively related to the strong-robustness.

Table 9: Accuracy of the VGG model (Simonyan & Zisserman, 2014) obtained from two noise-perturbed testing cases in CIFAR-10 dataset (Krizhevsky & Hinton, 2009). The second column shows the result on randomly perturbed samples, and the third column shows the result on adversarially perturbed samples.

| Attack power (defined in Eq. (12.6)) | Test accuracy on randomly perturbed samples | Test accuracy on adversarially perturbed samples |
|---|---|---|
| 0 | 0.9395 | 0.9395 |
| 1 | 0.938 | **0.7807** |
| 5 | 0.938 | **0.3767** |
| 10 | 0.9377 | **0.2092** |

**Theorem 12.3.** $f_1$ *is strong-robust against adversarial examples if and only if* $ARC(f_1)/\delta_2 = 1$.

**Proof of Theorem (12.3):**

*Proof.* If $ARC(f_1)/\delta_2 = 1$, then based on Definition (12.1), we have that

$\mathbb{P}(d_2(x, x') = \delta_2) = 1$.

This indicates that

$\mathbb{P}(f_1(x) = f_1(x')|d_2(x, x') < \delta_2) = 1$, which is the exact definition of strong-robustness ( Eq. (4.8)).

If $f_1$ is strong-robust, then $\mathbb{P}(f_1(x) = f_1(x')|d_2(x, x') < \delta_2) = 1$.

Therefore $ARC(f_1) = \mathbb{E}[d_2(x, x')]$. Since $\mathbb{P}(f_1(x) \neq f_1(x')|d_2(x, x') < \delta_2) = 0$, we have that

$$\begin{aligned}
ARC(f_1) &= \\
\mathbb{E}[d_2(x, x')] &= \\
\delta_2\mathbb{P}(f_1(x) = f_1(x')|d_2(x, x') < \delta_2) &= \\
\delta_2
\end{aligned} \tag{12.4}$$

$ARC(f_1)/\delta_2 = 1$. □

### 12.3 Using "Siamese Architecture" to improve DNNs' Adversarial Robustness

One intuitive formulation that we can use to improve a DNN's adversarial robustness is by solving the following:

$$\begin{aligned}
\forall x, x' \in X, &\text{if } d_2(g_2(x), g_2(x')) < \epsilon, \\
&\underset{w}{\mathrm{argmin}}\, d_1(g_1(x; w), g_1(x'; w))
\end{aligned} \tag{12.5}$$

This essentially forces the DNN to have the finer topology between $(X_1, d_1)$ and $(X_2, d_2)$ by learning a better $g_1$. We name the strategy minimizing the loss defined in Eq. (12.5) as "Siamese Training" because this formulation uses the Siamese architecture (Bromley et al., 1993), a classical deep-learning approach proposed for learning embedding. We feed a slightly perturbed input $x'$ together with its original seed $x$ to the Siamese network which contains two copies (sharing the same weights) of a DNN model we want to improve. By penalizing the difference between middle-layer ($g_1(\cdot)$) outputs of $(x, x')$, "Siamese Training" can push two spaces $(X, d_1')$ versus $(X_2, d_2')$ to approach finer topology relationship, and thus increase the robustness of the model. This can be concluded from Figure 6. By assuming $d_2(g_2(x), g_2(x'))$ equals (approximately) to $||\Delta(x, x')||$, previous studies (summarized in Table 2) normally assume $d_2$ is a norm function $|| \cdot ||$. Because for a pair of inputs $(x, x')$ that are close to each other (i.e., $||x - x'||$ is small) in $(X, || \cdot ||)$, Siamese training pushes them to be close also in $(X_1, d_1)$ . As a result, this means that a sphere in $(X_1, d_1)$ maps to a not-too-thin high-dimensional ellipsoid in $(X, || \cdot ||)$. Therefore the adversarial robustness of DNN model after Siamese training may improve. In experiments, we choose Euclidean distance $|| \cdot ||_2$ for $d_1(\cdot)$ (however, many other choices are possible).

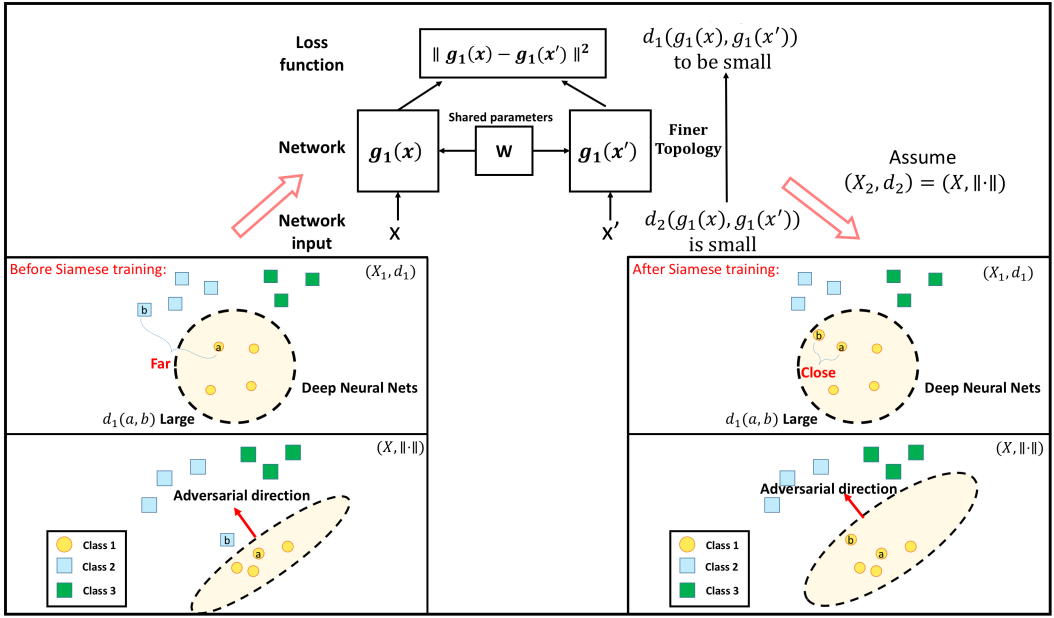

Figure 6: Sketch of Siamese training. Inputs are pairs of seed sample and their randomly perturbed version, while we suppose the $d_2$ distance between the pair is small. By forwarding a pair into the Siamese network and penalizing the outputs of the pair, this training intuitively limit the $d_1$ distance between two similar samples to be small. Backpropagation is used to update the weights of the network.

## 12.4 COMPARING DNN HARDENING STRATEGIES EXPERIMENTALLY

**Datasets:** Currently, we are using the following 2 image datasets to evaluate our model:

- **MNIST:** MNIST, released in (LeCun et al., 1998) includes a task to classify handwritten digits. It has a training set of 60,000 examples, and a test set of 10,000 examples. Each example is a 32x32 pixel black and white image of handwritten digit.
- **CIFAR-10:** CIFAR-10 is an image classification dataset released by (Krizhevsky & Hinton, 2009). The training set contains 50,000 32x32 color images in 10 classes, and the test set contains 10,000 32x32 color images. **VGG model:** We choose a VGG model (Simonyan & Zisserman, 2014) as a base DNN model. The VGG model in our experiment has 16 weight layers (55 layers in total).

**Baseline:** Three different hardening strategies are compared through testing on adversarial examples (details in Section 6.2.1): (1) original model; (2) stability training (Zheng et al., 2016) [18]; (3) Siamese training (alone); (4) adversarial training (Goodfellow et al., 2014; Kurakin et al., 2016) uses adversarial perturbed $x'$ and original samples $x$ to train a DNN model.

The first column of Table 10 and Table 11 shows different levels of attack power (defined in Eq. (12.6)). Test accuracy reported in Figure 7(a), Figure 8(a), Table 10 and Table 11 shows different hardening approches can increase the effectiveness of the adversarial attacks. Details of our experimental set-up and datasets are included in Section 6.2.1.

**Evaluation Metrics:**

---

[18]ATT: Stability training was shown to improve the model robustness against Gaussian noise in (Zheng et al., 2016). Differently, our experiments focus on testing a learning model's robustness against "adversarial perturbation". The sole purpose of including this baseline is to show where state-of-art hardening strategies are in our experimental setting.

Table 10: Test accuracy for different training strategies on CIFAR-10 dataset.

| Attack power (Eq. (12.6)) | Original model | Stability Training | Siamese Training |
|---|---|---|---|
| 0 | 93.95% | 93.81% | 93.96% |
| 1 | 78.07% | 78.01% | 93.88% |
| 2 | 61.38% | 60.34% | 90.13% |
| 3 | 50.07% | 49.21% | 86.73% |
| 4 | 42.86% | 41.51% | 83.85% |
| 5 | 37.67% | 36.33% | 81.21% |
| 6 | 33.60% | 32.08% | 78.61% |
| 7 | 29.70% | 28.09% | 76.09% |
| 8 | 26.23% | 25.11% | 73.21% |
| 9 | 23.53% | 22.43% | 69.67% |
| 10 | 20.92% | 20.25% | 65.98% |
| ARC | 4.9798 | 4.8717 | **8.9332** |
| ARCA | 0.4253 | 0.4155 | **0.7631** |

Table 11: Test accuracy for different training strategies on MNIST dataset.

| Attack power | Original model | Adversarial Training | Stability Training | Siamese Training |
|---|---|---|---|---|
| 0 | 98.98% | 98.96% | 99.06% | 99.03% |
| 1 | 98.75% | 98.84% | 98.94% | 98.84% |
| 2 | 98.44% | 98.63% | 98.60% | 98.47% |
| 3 | 98.10% | 98.41% | 98.29% | 98.16% |
| 4 | 97.56% | 98.12% | 97.80% | 97.78% |
| 5 | 97.09% | 97.80% | 97.47% | 97.26% |
| 6 | 96.23% | 97.38% | 97.01% | 96.56% |
| 7 | 95.43% | 96.96% | 96.23% | 95.81% |
| 8 | 94.22% | 96.47% | 95.37% | 95.01% |
| 9 | 92.95% | 96.06% | 94.49% | 93.89% |
| 10 | 91.53% | 95.57% | 93.30% | 92.76% |
| ARC | 10.5928 | 10.732 | 10.6656 | 10.6357 |
| ARCA | 0.953159 | 0.96549 | 0.960486 | 0.957503 |

- **Test accuracy:** We use top-1 test accuracy as the performance metric. It is defined as the number of successfully classified samples divided by the number of all test samples. The base model achieves accuracy when there's no adversarial attack.
- **ARC (Eq. (12.2)) :** We use ARC to measure the adversarial robustness of each model. $\eta$ is chosen to be 10.
- **ARCA: (Eq. (12.3)) :** We use ARCA to measure the total performance of each model.

We generate adversarial examples using the fast gradient sign method, in which the power of the adversary attack can be easily controlled. By controlling the power of fast-sign attacks, we can obtain a complete view of how the accuracy changes according to different attack powers.

In the following analysis, the attack power is defined as:
$$P = ||x - x'||_\infty \tag{12.6}$$
For image classification tasks, we control the perturbed sample to be still in the valid input space, so that every dimension of the perturbed samples is in the range of integers between 0 and 255.

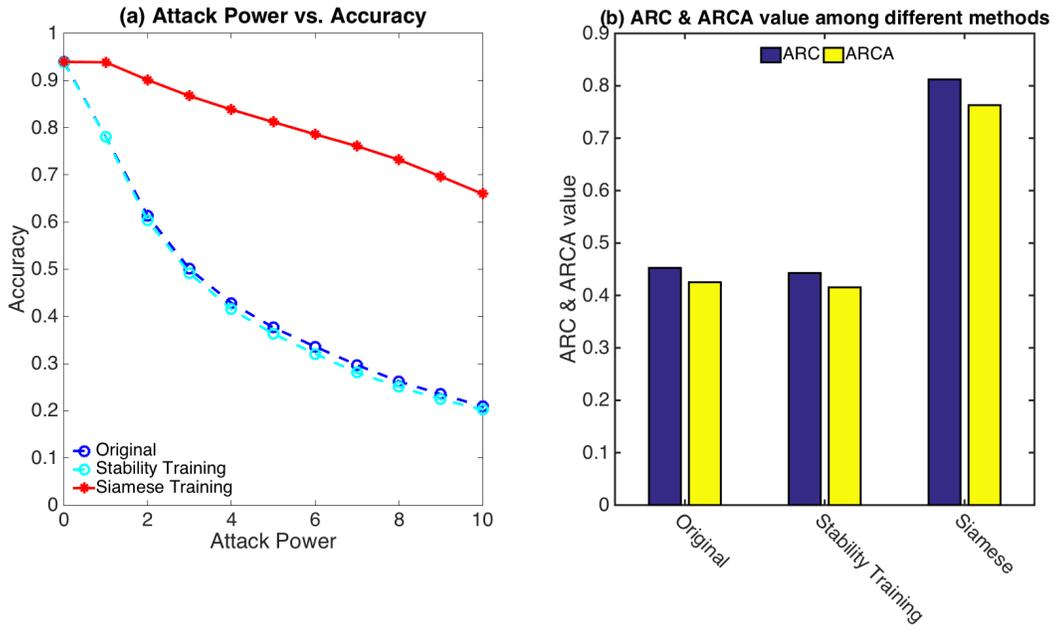

Figure 7: Result of CIFAR-10: (a) Test accuracy under adversarial example attacks: three different colors for three different training strategies. (Details in Section 6.2.1) We don't include the result of adversarial training because previous adversarial training can't be used on networks with batch-normalization. Some tricks of training such networks are released in a recent paper (Kurakin et al., 2016) (b) ARC and ARCA for three different training strategies under adversarial example attacks.

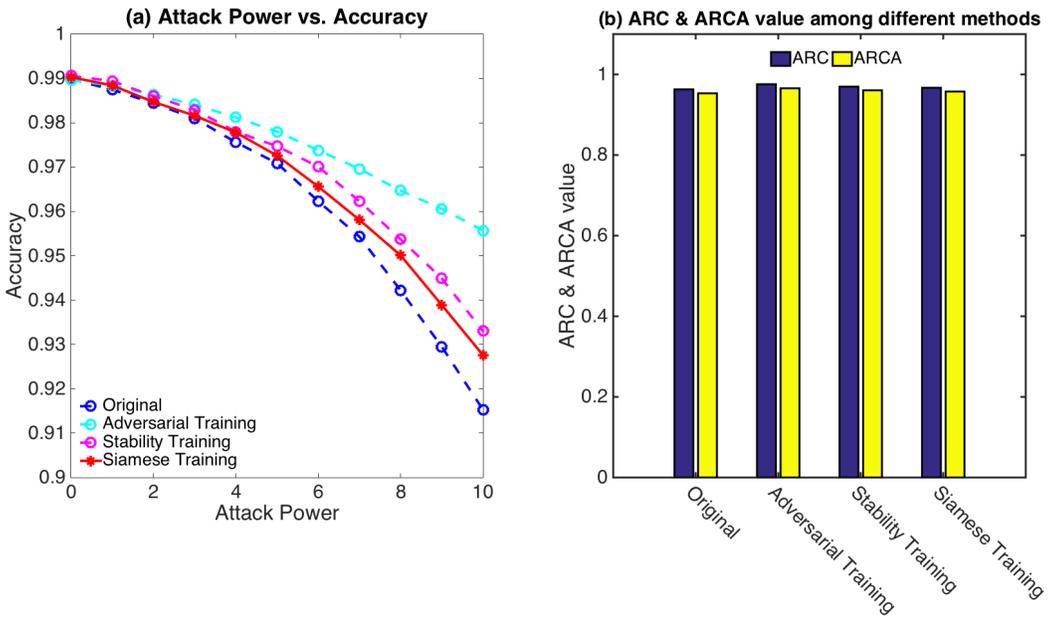

Figure 8: (a) Test accuracy under adversarial example attacks on MNIST dataset: four different colors for four different training strategies. (Details in Section 6.2.1) (b) ARC and ARCA for four different training strategies under adversarial example attacks.

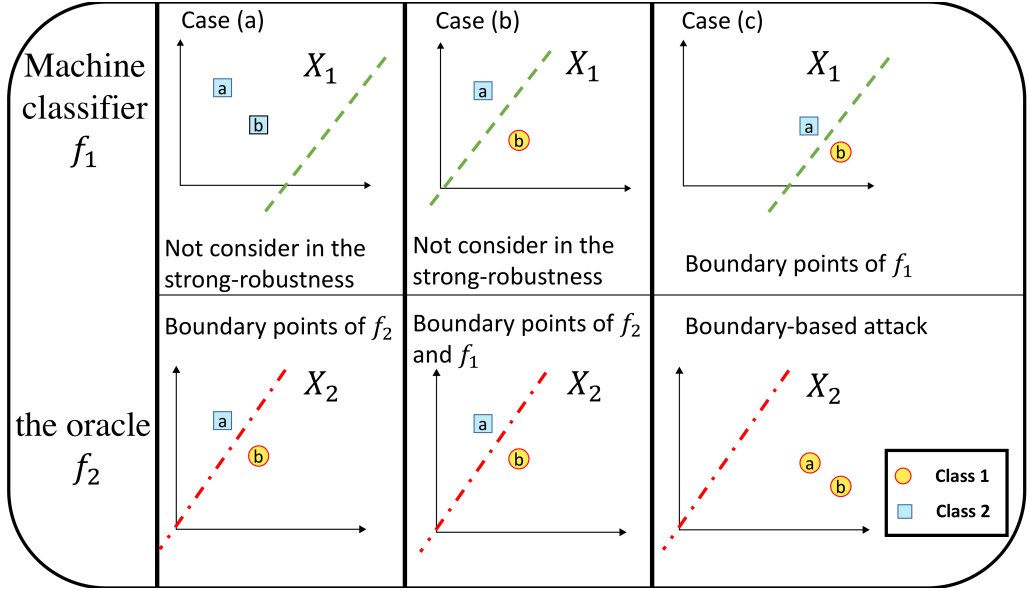

Figure 9: An example showing boundary points of $f_1$ and boundary points of $f_2$. We assume $f_1$ and $f_2$ are continuous a.e.. We assume $c_1$ and $c_2$ as linear classification functions. The first two columns showing boundary points of $f_2$ that are not considered in this paper. The third column describes "Boundary based adversarial attacks" that can only attack seed samples whose distance to the boundary of $f_1$ is smaller than $\epsilon$. Essentially this attack is about those boundary points of $f_1$ that are treated as similar and belong to the same class by $f_2$.

## 13 Boundary Points of $f_1$ Matter for adversarial examples When $f_1$ is Not Continuous a.e.

When $f_1$ is not continuous a.e., the analysis of adversarial examples needs to consider "boundary points" of $f_1$ with certain properties. This section tries to clarify the definition and related scope.

**Definition 13.1.** *We define the set of boundary points of $f_i$ as the following set of sample pairs:*
$$\{(x, x') | f_i(x) \neq f_i(x'), d_i(g_i(x), g_i(x')) < \delta_i, \\ x \in X, x' \in X\} \tag{13.1}$$

Our definition of the boundary points describes such points as pairs of samples that are across the classification boundary. This format of definition makes the following analysis (notation-wise) easy and concise.

**Lemma 13.2.** $f_i$ *is not continuous a.e., if and only if*
$$x \in X, x' \in X, \mathbb{P}(f_i(x) \neq f_i(x') | d_i(g_i(x), g_i(x')) < \delta_i) > 0 \tag{13.2}$$

This lemma shows that a case with probability of boundary points larger than 0 is exactly the situation when $f_i$ being not continuous a.e..

### 13.1 More about boundary points of $f_1$ and boundary points of $f_2$

In addition, we want to point out that all boundary pairs of $f_2$ (satisfying $f_2(x) \neq f_2(x')$ and $d_2(g_2(x), g_2(x')) < \delta_2$) are not considered in our analysis of adversarial examples. Figure 9 illustrates three types of boundary points, using the first two columns showing boundary points of $f_2$.

The third column of Figure 9 describes "Boundary based adversarial examples" that can only attack seed samples whose distance to the boundary of $f_1$ is smaller than $\delta_2$. Essentially this attack is about

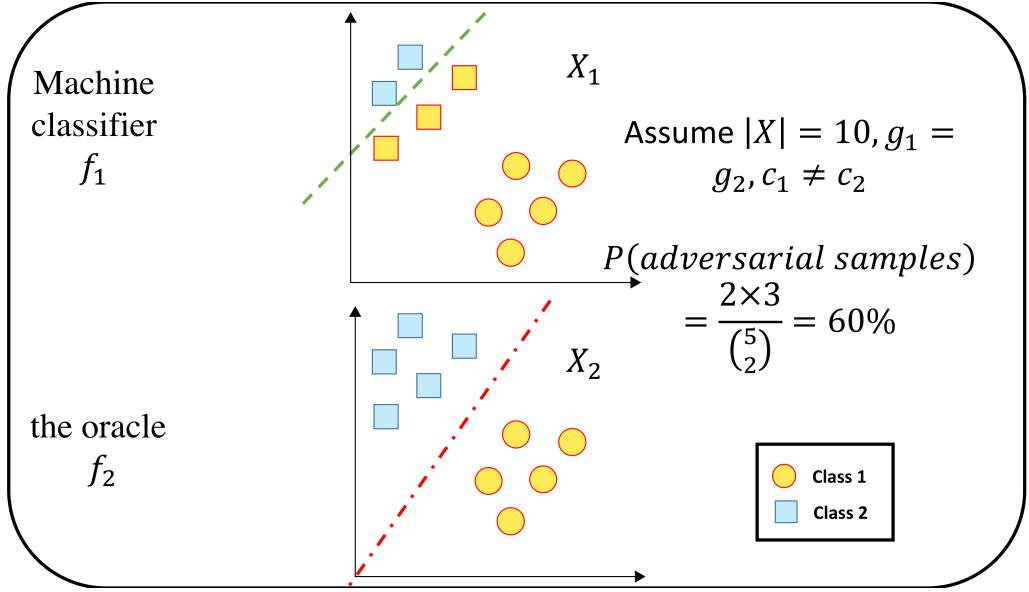

Figure 10: When $f_1$ is not continuous a.e., the strong-robustness of $f_1$ is determined by both $g_1$ and $c_1$. We assume $c_1$ and $c_2$ as linear classification functions. This figure shows when (1) sample space $X$ is finite, (2) $f_1$ learns a wrong decision boundary and (3) the probability of test samples around $f_1$'s decision boundary is large, $f_1$ is not strong-robust against adversarial examples. However, we want to emphasize that the situation is very rare for a well-trained classifier $f_1$.

those boundary points of $f_1$ that are treated as similar and belong to the same class by $f_2$. That is

$$\mathbb{P}(f_1(x) \neq f_1(x')|f_2(x) = f_2(x'), d_2(g_2(x), g_2(x')) < \delta_2,$$
$$d_1(g_1(x), g_1(x')) < \delta_1) \quad (13.3)$$

- When $f_1$ is continuous a.e., Eq. (4.1) equals to 0. (derived from Eq. (8.1) in Section 8)
- When $f_1$ is not continuous a.e., Eq. (4.1) might be larger than 0. (derived from Eq. (13.2))

The value of this probability is critical for our analysis in Theorem (4.3) and in Theorem (4.5). Again, we want to emphasize that most machine learning methods assume $f_1$ is continuous a.e. and therefore "boundary based adversarial attacks" are not crucial.

### 13.2 When $f_1$ not Continuous a.e., Strong-robust is significantly influenced by boundary points of $f_1$

When $f_1$ is not continuous a.e., for instance when $X$ is a finite space, the probability of "adversarial examples" can be calculated as:

$$\mathbb{P}(f_1(x) \neq f_1(x')|f_2(x) = f_2(x'), d_2(g_2(x), g_2(x')) < \delta_2)$$
$$= \frac{\#\{(x, x')|f_2(x) = f_2(x') \& d_2(g_2(x), g_2(x')) < \delta_2 \& f_1(x) \neq f_1(x')\}}{\#\{(x, x')|f_2(x) = f_2(x') \& d_2(g_2(x), g_2(x')) < \delta_2\}} \quad (13.4)$$

This is exactly the proportion of those pairs of points for which $f_1$ classifies them into different classes and $f_2$ treats them as similar and "same-class" samples. For this case, both $g_1$ and $c_1$ matter for the strong-robustness of $f_1$. See Appendix Section 13.2 for an example showing how $c_1$ makes $f_1$ not strong robust.

#### 13.2.1 $c_1$ matters for strong-robustness when $f_1$ is not a.e. continuous

Based on Eq. (13.4), when $f_1$ is not continuous a.e., the strong-robustness of $f_1$ is determined by both $g_1$ and $c_1$. Figure 10 shows an exemplar case in which $X$ has only ten samples (i.e. $|X| = 10$).

We assume the learned $f_1$ and the oracle $f_2$ derive the same feature space, i.e., $X_1 = X_2$. And we also assume $f_1$ performs the classification very badly because the decision boundary (by $c_1$) on $X_1$ is largely different from the decision boundary on $X_2$. The probability of "adversarial examples" in this case can be calculated by using Eq. (13.4). We get $\mathbb{P}(f_1(x) \neq f_1(x')|f_2(x) = f_2(x'), d_1(g_1(x), g_1(x')) < \delta_1) = \frac{2*3}{5*2} = 0.6$.

Clearly in this case, $c_1$ matters for the strong-robustness (when $f_1$ is not a.e. continuous). This figure indicates that when (1) sample space $X$ is finite, (2) $f_1$ learns a wrong decision boundary and (3) the probability of test samples around $f_1$'s decision boundary is large, $f_1$ is not strong-robust against adversarial examples. However, we want to point out that this situation is very rare for a well-trained classifier $f_1$.

For cases when $f_1$ is not continuous a.e., obtaining more samples is clearly a good way to learn a better decision boundary that might improve the adversarial robustness of the classifier at the same time.